\newif\ifconfver
\confverfalse      

\newif\ifcutshort      
\cutshorttrue

\newif\ifcutshortlvltwo  
\cutshortlvltwofalse

\ifconfver
\documentclass[10pt,twocolumn,twoside]{IEEEtran}
\else
\documentclass[12pt, draftclsnofoot, onecolumn]{IEEEtran}
\fi

\IEEEoverridecommandlockouts

\usepackage{titlesec}
\usepackage{amsmath,graphicx}
\usepackage{booktabs}
\usepackage{textcomp}
\usepackage{graphicx}
\usepackage{psfrag}
\usepackage{subfigure}
\usepackage{url}
\usepackage{stfloats}
\usepackage{amsfonts,amssymb,amsmath,amsbsy,bm,theorem,ifthen,color}
\usepackage{array}
\usepackage{arydshln}
\usepackage{calc}
\usepackage{subfig}
\usepackage{longtable}
\usepackage{multirow}
\usepackage{algorithmic,algorithm}
\usepackage{graphics,booktabs,epsfig,enumerate}
\usepackage{afterpage}
\usepackage{float}
\usepackage{cite}
\usepackage{xcolor}
\usepackage{threeparttable}
\usepackage{todonotes}
\usepackage{lipsum}
\usepackage{verbatim}
\usepackage{mathtools}
\usepackage{colortbl}

{\begin{list}{}{
		\settowidth{\labelwidth}{\mbox{\textnormal{#1}}}%
		\setlength{\leftmargin}{\labelwidth+\labelsep}}}%
{\end{list}}

\newcommand\Ec{\ensuremath{\mathcal{E}}}

\newcommand\Nc{\ensuremath{\mathcal{N}}}
\newcommand\Xc{\ensuremath{\mathcal{X}}}

\newcommand\Dc{\ensuremath{\mathcal{D}}}

\newcommand\Sc{\ensuremath{\mathcal{S}}}

\newcommand\Oc{\ensuremath{{\mathcal{O}}}}

\newcommand\xb{\ensuremath{{\bm x}}}
\newcommand\yb{\ensuremath{{\bm y}}}

\newcommand\wb{\ensuremath{{\bm w}}}

\newcommand\ellb{\ensuremath{{\bm {\ell}}}}

\newcommand\xib{\ensuremath{{\bm \xi}}}

\newcommand\zerob{\ensuremath{{\bm 0}}}

\newcommand{\wt}{\widetilde}

\newtheorem{Lemma}{Lemma}

\newtheorem{Theorem}{Theorem}
\newtheorem{Def}{Definition}
\newtheorem{Corollary}{Corollary}

\theorembodyfont{\rmfamily}

\newtheorem{Remark}{Remark}
\definecolor{orange}{RGB}{255,107,0}

\def\BibTeX{{\rm B\kern-.05em{\sc i\kern-.025em b}\kern-.08em
    T\kern-.1667em\lower.7ex\hbox{E}\kern-.125emX}}
\allowdisplaybreaks[4]
\begin{document}
\title{Federated Learning with Enhanced Privacy via Model Splitting and Random Client Participation}

\author{Yiwei~Li,~\IEEEmembership{Senior Member,~IEEE},
 ~Shuai~Wang,
~Zhuojun~Tian,
~Xiuhua~Wang,
~Shijian~Su,~\IEEEmembership{Senior Member,~IEEE}
\thanks{The work of Yiwei Li was supported by the Natural Science Foundation of Xiamen City under Grant 3502Z202573070, and by the Key Science and Technology Project of Xiamen City under Grant 3502Z20251011.
The work of Shuai Wang was supported in part by the Fundamental Research Funds for the Central Universities, under Grant ZYGX2024XJ072, and in part by the National Natural Science Foundation of China, under Grant 62401121.    }
\IEEEcompsocitemizethanks{\IEEEcompsocthanksitem Y. ~Li is with the School of Optoelectronic \& Communication Engineering, Xiamen University of Technology, Xiamen, 361024, China (e-mail: lywei0306@foxmail.com).\protect
\IEEEcompsocthanksitem S. ~Wang is with the National Key Laboratory of Wireless Communications, University of Electronic Science and Technology of China, Chengdu, 611731, China (e-mail: shuaiwang@uestc.edu.cn). \protect
\IEEEcompsocthanksitem   Z.~Tian is with the Division of Information Science and Engineering, KTH Royal Institute of Technology, Stockholm, Sweden. (e-mail: zhuojun@kth.se).  \protect
\IEEEcompsocthanksitem   X.~Wang is with the School of Cyber Science and Engineering, Huazhong University of Science and Technology, Wuhan, 430074, China (e-mail: xiuhuawang@hust.edu.cn).  \protect
\IEEEcompsocthanksitem   S.~Su is with  the School of Engineering, Huaqiao University, Quanzhou, 362021, China (e-mail: shijiansu@foxmail.com).  \protect
 }}

\maketitle
\begin{abstract}	
Federated Learning (FL) often adopts differential privacy (DP) to protect client data, but the added noise required for privacy guarantees can substantially degrade model accuracy. To resolve this challenge, we propose model-splitting privacy-amplified federated learning (MS-PAFL), a novel framework that combines structural model splitting with statistical privacy amplification. In this framework, each client’s model is partitioned into a private submodel, retained locally, and a public submodel, released for global aggregation. The calibrated Gaussian noise is injected only into the public submodel, thereby confining its adverse impact while preserving the utility of the retained local model components.
We further present a rigorous theoretical analysis that characterizes the joint privacy amplification achieved through random client participation and local data subsampling under this architecture. The analysis provides explicit bounds on both single-round and total privacy loss, demonstrating that MS-PAFL significantly reduces the noise necessary to satisfy a target privacy protection level. Extensive experiments validate our theoretical findings, showing that MS-PAFL consistently attains a superior privacy--utility trade-off compared with conventional DP-FL baselines and enables the training of highly accurate models under strong privacy guarantees.

\vspace{0.05cm}
\\\\
\end{abstract}

\section{Introduction}\label{sec:Introduction}
\IEEEPARstart{F}{ederated} Learning (FL) has emerged as a powerful distributed machine learning (ML) paradigm, enabling collaborative model training across a large number of clients under the coordination of a central parameter server (PS), without requiring direct access to their raw data~\cite{ma2020safeguarding,li2025differentially}. Although FL's decentralized nature offers inherent privacy benefits, it remains vulnerable to privacy leakage.  Sophisticated adversaries can potentially infer sensitive information by analyzing the sequence of model parameters exchanged between clients and the PS~\cite{zhang2021fedpd,li2019convergence, triastcyn2019federated,10643038}.
Differential Privacy (DP) has therefore become  the \textit{de facto} standard for providing formal privacy guarantees in FL, typically by injecting calibrated noise into communicated model updates~\cite{li2024privacy,li2024diffprivate}.  This mechanism effectively obscures the contribution of any single client, making it computationally infeasible for an adversary to determine whether an individual's data participated in the model training. However, stronger privacy protection requires stronger perturbation, which often slows convergence and degrades model accuracy. This privacy--utility trade-off remains one of the main bottlenecks in practical DP-FL design~\cite{11017481,10032626,10179278}.

To alleviate the performance degradation caused by DP noise, privacy amplification has received considerable attention~\cite{li2022network,girgis2021shuffled,feldman2018privacy}.   This statistical technique leverages stochasticity in the training process to strengthen privacy guarantees by increasing uncertainty about the participation of individual clients or data samples. In FL, privacy amplification is mainly realized through client sampling and data subsampling~\cite{balle2018privacy,steinke2022composition}.
These sampling strategies introduce uncertainty regarding whether any specific client or data point contributes to a given update. This uncertainty is the essence of the ``amplification”: it forces an adversary to consider an exponentially larger hypothesis space, thereby diluting potential information leakage and enabling the same privacy level with less noise.  While existing studies~\cite{erlingsson2019amplification,10387292,10049722,10959094} have analyzed these mechanisms separately, their joint impact remains an open and challenging research question. More importantly, existing statistical methods only reduce the amount of noise required but do not address its indiscriminate application across the entire model vector, which may corrupt parameters critical to model performance.

Although privacy amplification provides  a powerful statistical means to mitigate the cost of DP, it does not address a more fundamental structural limitation of typical DP-FL frameworks. The prevailing strategy of injecting noise into the entire model update vector is indiscriminate: it overlooks the fact that some parameters are naturally robust to perturbation, whereas others are critical for maintaining accuracy~\cite{11164482,10740400}. As a result, even when privacy amplification reduces the required noise magnitude, uniform corruption across all parameters can still severely degrade performance. This limitation motivates a shift in perspective---from merely reducing the amount of noise to optimizing where the noise is applied.
To this end, we consider partitioning each client model into two components: a private submodel that remains local and a public submodel that is released for global aggregation. DP noise is then injected only into the released public submodel, thereby reducing distortion on the retained private component. This design introduces a structural mechanism that complements statistical privacy amplification. Rather than relying solely on reducing the magnitude of injected noise, it preserves model fidelity through a principled structural design. More importantly, we argue that an effective privacy--utility trade-off is best achieved through the synergistic integration of structural and statistical mechanisms, which enables stronger privacy protection while incurring only limited degradation in learning performance.
We also note that, in practical wireless edge FL systems, system-level factors such as sensing, communication, and computation may further influence learning performance and overall system efficiency~\cite{wen2025integrated}. The present work focuses on characterizing which model components are retained locally, which components are released for aggregation, and how client sampling and data subsampling jointly affect the privacy loss of the released public submodel.

\subsection{Related works}
\subsubsection{Differentially Private Federated Learning}
The DP-FL framework has become a standard approach for providing formal privacy guarantees. The moments accountant method introduced in~\cite{abadi2016deep} for tracking total privacy loss in centralized deep learning was later extended to the federated setting by~\cite{mcmahan2017communication}, giving rise to the widely adopted DP-FedAvg algorithm~\cite{li2020secure,9069945}. In this approach, clients clip their model updates and inject  calibrated noise before transmitting them to the PS for model aggregation. While this method achieves a strong privacy guarantee, it suffers from a persistent limitation: the injected noise often slows convergence and degrades model accuracy, resulting in a challenging privacy--utility trade-off~\cite{kairouz2021advances,li2019convergence}. Several works~\cite{9005465,triastcyn2019federated,truex2020ldp} have attempted to mitigate this degradation, but the fundamental challenge of global noise injection remains a critical bottleneck.

Along this line, recent studies have moved beyond the standard homogeneous DP-FL setting and explored more refined privacy designs. Dynamic personalized DP-FL methods exploit adaptive parameter retention to reduce the impact of clipping and noise on informative local parameters~\cite{yang2023dynamic}. Noise-aware heterogeneous DP-FL methods improve aggregation under non-uniform privacy/noise levels across clients~\cite{malekmohammadi2024noise}. Across-silo user-level DP has also been investigated to provide stronger privacy guarantees in more complex federated settings~\cite{kato2024uldp}. More recently, time-adaptive privacy spending has been proposed to allocate privacy budgets non-uniformly across rounds and clients for improved privacy--utility trade-offs~\cite{kiani2025differentially}.  However, most existing studies still treat the communicated model update as a monolithic object, and mainly improve utility through better clipping, aggregation, or accounting. As a result, the potential role of structural parameter separation in formal DP design remains relatively less explored in the current DP-FL literature.

\subsubsection{Privacy Amplification via Sampling}
Privacy amplification has emerged as a valuable statistical tool for mitigating the utility loss of DP~\cite{feldman2023stronger,erlingsson2019amplification,choquette2023privacy}. The key idea is that introducing randomness into the data processing pipeline strengthens privacy guarantees, enabling the same level of privacy protection with reduced noise. In FL, two main sampling schemes have been studied. The first is data subsampling, typically implemented via stochastic gradient descent (SGD) on mini-batch data, which enhances privacy by making the contribution of each individual data point uncertain~\cite{balle2018privacy,steinke2022composition,fang2024privacy,feldman2025privacy}.  The second scheme is client sampling, where only a random subset of clients participates in each round, further amplifying privacy. The work in~\cite{balle2020privacy} formally analyzed this ``privacy amplification via random check-ins,'' showing that uncertainty about clients' participation in a given round yields meaningful privacy benefits.
Overall, these studies establish privacy amplification as an effective statistical tool in DP-FL. However, most existing analyses focus on data subsampling and client sampling separately, or treat sampling as an auxiliary component within a conventional full-model perturbation pipeline. As a result, the joint amplification effect of client participation and local data subsampling in FL is still not fully characterized, especially when sampling interacts with more structured model designs beyond standard full-model updates.

\subsubsection{Structural Privacy Mechanisms and Model Splitting}
Beyond statistical methods, recent works have also explored structural modifications to the FL protocol to enhance privacy or personalization. One notable example is split learning~\cite{vepakomma2018split,thapa2022splitfed,zhang2023privacy}, where the neural network is vertically partitioned between the client and the PS~\cite{10938701}. While this avoids sharing raw data, it incurs high communication costs and does not inherently provide formal DP guarantees without additional mechanisms~\cite{10534285}. More recent hybrid frameworks such as privacy-preserving split federated learning have further combined FL and split learning to better address privacy and heterogeneity in edge environments~\cite{zheng2024ppsfl}.
Another highly related line is personalized federated learning (PFL), where model parameters are divided into shared/global and personalized/local parts to better fit non-identically and independently distributed (non-iid) data. Recent methods have shown that this parameter partitioning perspective is highly effective for personalization, including parameter propagation~\cite{wu2023personalized}, dynamic Fisher-based personalization under DP~\cite{yang2023dynamic}, customized parameter selection and subnetwork personalization~\cite{tamirisa2024fedselect}, and hypernetwork-based generation of personalized client models~\cite{scott2024pefll}.
Although structural partitioning has been widely recognized as a useful design tool in FL, its role in formal DP analysis remains much less understood. In particular, existing studies have not clearly characterized how structural parameter separation interacts with the joint privacy benefits induced by client sampling and local data subsampling.

Overall, recent literature suggests that both privacy accounting and parameter partitioning are important for practical FL systems. Nevertheless, to the best of our knowledge, existing works have not provided a unified framework that simultaneously combines: 1) model splitting into retained private and released public components, 2) selective DP perturbation applied only to the released public submodel, and 3) rigorous privacy amplification analysis under joint random client participation and local data subsampling. This gap motivates the proposed model-splitting privacy-amplified federated learning (MS-PAFL) framework.

\subsection{Contributions}
In this paper, we propose MS-PAFL, a novel framework that integrates structural and statistical mechanisms to enhance DP in FL systems. Specifically, MS-PAFL partitions each client’s model into a private submodel and a public submodel,  and injects calibrated noise only into the latter, thereby reducing the impact of perturbation on the retained model components. Combined with client participation and data subsampling, this design achieves a stronger balance between privacy and utility. The main contributions are as follows:

\begin{itemize}
\item \textbf{Hybrid framework:} We introduce a unified FL framework that combines model splitting with privacy amplification via random client participation and local data subsampling, offering a more effective solution to the privacy--utility trade-off.

\item \textbf{Theoretical guarantees:} We theoretically analyze single-round and total privacy loss bounds, analytically showing how splitting and sampling jointly strengthen privacy. Our analysis explicitly isolates the privacy loss to the public submodel, which is the only communicated component under MS-PAFL.

\item \textbf{Empirical validation:} Extensive experiments confirm that MS-PAFL achieves superior privacy--utility trade-offs, consistently outperforming standard DP-FL baselines.  We further provide ablations on the splitting ratio and sampling parameters, as well as evaluations under broader settings, to validate the benefit of combining structural partitioning with statistical amplification.
\end{itemize}

{\bf Synopsis:} For ease of the ensuing presentation, the main mathematical notations used are listed in Table \ref{tab:Notations}. Section~\ref{sec:Preliminaries} introduces the preliminaries of FL and DP. Section \ref{sec:MS-PAFL} presents the proposed MS-PAFL framework and its components. Section \ref{subsec:privacy_analysis} presents our comprehensive privacy analysis. The experimental results are presented in Section \ref{sec:Experimental Details}, and conclusions are drawn in Section~\ref{sec:Conclusions}.

\begin{table}[t]
\renewcommand{\arraystretch}{1.25}
\centering
\caption{Notations}
\begin{tabular}{m{2.1cm}|m{5.9cm}}
\hline
\hline
\textbf{Notation} &  \textbf{Description} \\
\hline
${\mathbb R}^d$ & The set of real $d\times 1$ vectors.\\
\cdashline{1-2}[0.8pt/1pt]
$[N]$ & The integer client set $\{1,\ldots, N\}$.\\
\cdashline{1-2}[0.8pt/1pt]
$|\Dc|$ & The size of set $\Dc$.\\
\cdashline{1-2}[0.8pt/1pt]
$b$ & The mini-batch data size.\\
\cdashline{1-2}[0.8pt/1pt]
$Q$ & The number of local SGD steps.\\
\cdashline{1-2}[0.8pt/1pt]
$K$ & The number of participating clients in each round.\\
\cdashline{1-2}[0.8pt/1pt]
$T$ & The total number of communication rounds.\\
\cdashline{1-2}[0.8pt/1pt]
$d_p, d_\alpha$ &  Dimensions of the public and private submodels, with $d_p+d_\alpha=d$.\\
\cdashline{1-2}[0.8pt/1pt]
$p_i^t$ &  Check-in probability of client $i$ at round $t$.\\
\cdashline{1-2}[0.8pt/1pt]
$\Xc^t$ &  Random set of participating clients in round $t$.\\
\cdashline{1-2}[0.8pt/1pt]
${\xb}^{\top}$ & Transpose of vector ${\xb}$.\\
\cdashline{1-2}[0.8pt/1pt]
$\{\xb_i \}$ & Set of $\xb_1,\xb_2,\dots$ for all the admissible $i$.\\
\cdashline{1-2}[0.8pt/1pt]
${\bf I}_d$ & The $d\times d$ identity matrix.\\
\cdashline{1-2}[0.8pt/1pt]
$\mathbb{Z}^{+}$ & The positive integer set.\\
\cdashline{1-2}[0.8pt/1pt]
$\|{\bf x}\|$ & The Euclidean norm of vector ${\bf x}$.\\
\cdashline{1-2}[0.8pt/1pt]
$\nabla f( \xb )$ & The gradient of function $f( \xb )$ with respect to $\xb$.\\
\cdashline{1-2}[0.8pt/1pt]
$\mathbb{E}[\cdot]$, $\mathbb{P} [\cdot]$ & The statistical expectation and the probability function, respectively.\\
\hline
\hline
\end{tabular}
\label{tab:Notations}
\end{table}

\section{Preliminaries}\label{sec:Preliminaries}
In this section, we briefly present the preliminaries of FL and DP, as well as the privacy amplification techniques relevant to the proposed framework.
\subsection{Preliminaries of FL}\label{subsec:FedAvg}
Let us consider a vanilla FL system that consists of one PS and $N$ clients. Suppose that each client $i \in [N]$ holds a private local dataset $\mathcal{D}_{i}\triangleq \big\{(\xb_{i,j}, \yb_{i,j}) \big\}_{j=1}^{|\mathcal{D}_{i}|}$ with size $|\mathcal{D}_{i}|$, where $\xb_{i,j}$ represents the feature of the $j$-th training data sample and $\yb_{i,j}$ is the corresponding label. The global objective function in this vanilla FL system can be formulated as follows:
\vspace{-0.2cm}
\begin{equation}\label{eq:EMR problem}
\min_{\wb}\Big\{ \frac{1}{N} \sum_{i=1}^{N}  F_{i}({\wb}) \Big\},
\end{equation}

\vspace{-0.1cm}
\noindent where $\wb \in \mathbb{R}^d$ denotes the global model parameters and $F_{i}: \mathbb{R}^{d} \rightarrow \mathbb{R}$ is the local objective function. Problem \eqref{eq:EMR problem} can be solved using the well-known algorithm, called FedAvg~\cite{li2020secure}.
To be specific, at the ${t}$-th communication round, FedAvg executes the following three steps:
\begin{enumerate}[(a)]
\item \textbf{Broadcasting}:
The PS randomly samples a subset of clients $\Sc^{t} \subseteq [N]$, and broadcasts global model $\overline \wb^{t-1}$ to all selected clients.
\vspace{0.1cm}
\item \textbf{Local update}: Each client $i \in \Sc^{t}$ updates local model by $Q$ steps of SGD, as follows
\begin{subequations}\label{eqn:local model}
\begin{align}
& \wb^{t,0}_{i}  =  \overline \wb^{t-1} \label{eqn:initiliazed_w}, \\
&  \wb_{i}^{t,r}   =  {\wb}^{t,r-1}_{i}  -   \eta_{t} \nabla F_{i}(\wb_{i}^{t,r-1}; \mathcal{B}_i^{t,r}) , r= 1, \ldots , Q,
\end{align}\label{eqn:local_Q_update}
\end{subequations}

\vspace{-0.3cm}
\noindent where $\wb_{i}^{t,r}$ is the local model of client $i$ at $r$-th inner iteration, $\eta_t$ is the learning rate, $\mathcal{B}_{i}^{t,r} \subseteq \Dc_i$ is a mini-batch dataset with $|\mathcal{B}_{i}^{t,r}| = b$ , and $\nabla F_{i}(\wb_{i}^{t,r-1}; \mathcal{B}_{i}^{t,r})$ is the mini-batch gradient. After $Q$ steps of local updates, we denote $\wb_i^{t} = \wb_i^{t,Q}$ for notational simplicity.

\vspace{0.1cm}
\item \textbf{Aggregation}: The selected clients upload their local models $\wb_i^{t,Q}, \forall i \in \Sc^{t}$ to the PS for the aggregation of global model $\overline\wb^{t}$, that is
\begin{align}
\overline\wb^{t} = \frac{1}{|\Sc^t|} \sum_{i \in \Sc^t}  \wb_i^{t}.
\end{align}
\end{enumerate}

\subsection{Preliminaries of DP}\label{subsec:Privacy Concern}
Unless otherwise stated, we consider an honest-but-curious threat model, where the PS follows the protocol but may attempt to infer private information from communicated messages. Clients may also be curious, but do not deviate from the prescribed protocol.  To address such privacy risks, a privacy-preserving mechanism should be employed to provide rigorous protection against these adversaries.
 The local DP (LDP) and central DP (CDP) are formally defined as follows:
\begin{Def}\label{Def:LDP}  $(\epsilon_{\ell}, \delta_{\ell})$-LDP {\rm \cite{balle2020privacy}}.
Suppose that $\Dc_{i}$ is the dataset of client $i$, a local randomized mechanism $\mathcal{M}_{i}: \mathcal{D}_{i} \rightarrow \mathbb{R}^{d}$ is $(\epsilon_{\ell}, \delta_{\ell})$-LDP if, for all pairs of neighboring datasets $\Dc_i$ and $\Dc_i^{\prime}$ that differ in one data sample and measurable subset $\Oc_{i} \subseteq Range(\mathcal{M}_{i})$, we have
\begin{align}\label{def LDP}
\mathbb{P}[\mathcal{M}_{i}(\Dc_i) \in \Oc_i] \leq \exp(\epsilon_{\ell}) \cdot \mathbb{P}\left[\mathcal{M}_{i}(\Dc_i^{\prime}) \in \Oc_i\right]+\delta_{\ell},
\end{align}
\end{Def}
where $\epsilon_{\ell}$ represents the privacy protection level of the local mechanism, and $\delta_{\ell} > 0$ indicates the probability of breaking the $(\epsilon_{\ell},0)$-LDP.  \noindent Here, $(\epsilon_{\ell},\delta_{\ell})$ denotes the per-round baseline DP guarantee of the Gaussian mechanism before amplification.

At the $t$-th round, the Gaussian mechanism requires carefully calibrated noise to be added to local models to guarantee $(\epsilon_{\ell}, \delta_{\ell})$-LDP, that is,
\begin{align}\label{eqn:DP_local_update}
\wt{\wb}_{i}^{t}  =   \wb_i^{t} + \boldsymbol{\xi}_i^t, ~\boldsymbol{\xi}_i^t \sim \Nc(\zerob, \sigma_{i,t}^{2}{\bf I}_d),
\end{align}where $\boldsymbol{\xi}_i^t$ is the artificial Gaussian noise with variance $\sigma_{i,t}^{2}$. The $\sigma_{i,t}^{2}$ required for achieving $(\epsilon_{\ell}, \delta_{\ell})$-LDP is given by the following lemma.
\begin{Lemma}  \label{Lemma: global sensitivity}
\cite[Theorem 3.22]{dwork2014algorithmic} Suppose a query function $g$ accesses the dataset $\Dc_i, \forall i \in [N]$ via randomized mechanism $\mathcal{M}$.
Let $\xi$ be zero-mean Gaussian noise with variance $\sigma^2$. Then, the minimal $\sigma^2$ required for $g+ \xi$ to satisfy $(\epsilon_{\ell},\delta_{\ell})$-DP is given by
\begin{align}
\sigma^2 = \frac{2 s^{2} \ln(1.25/\delta_{\ell})}{\epsilon_{\ell}^{2}},
\end{align}where $ s$ is the $\ell_2$-norm sensitivity of the function $g$ defined by
\begin{align}\label{eqn:global sensitivity_f}
s \triangleq \max _{\mathcal{D}_i, \mathcal{D}_i^{\prime}}\big\|g(\mathcal{D}_i)- g\big(\mathcal{D}_i^{\prime}\big)\big\|.
\end{align}
\end{Lemma}

Lemma~\ref{Lemma: global sensitivity} characterizes the noise level required for the local Gaussian mechanism to satisfy per-round LDP. However, in FL, the noisy local updates are further aggregated at the PS to form a released global model. Therefore, besides local privacy guarantees, we also need the following notion of central DP to characterize the privacy protection of the global model.

\begin{Def}\label{Def:CDP}  $(\epsilon_c, \delta_c)$-CDP {\rm \cite{balle2020privacy}}.
Let $\mathcal{D} \triangleq \mathcal{D}_{1} \times \mathcal{D}_{2} \times \cdots \times \mathcal{D}_{N}$ be the collection of all possible datasets from $N$ clients. Consider two neighboring datasets $\mathcal{D}$ and $\mathcal{D}^{\prime}$, which differ in only one data sample. A randomized mechanism $\mathcal{M}: \mathcal{D} \rightarrow \mathbb{R}^d$ is $(\epsilon_{c}, \delta_c)$-DP if for any two $\mathcal{D}$, $\mathcal{D}^{\prime}$ and measurable subset $\Oc \subseteq Range(\mathcal{M})$, we have
\begin{align}\label{def CDP}
\mathbb{P}[\mathcal{M}(\mathcal{D}) \in \Oc] \leq \exp(\epsilon_c) \cdot \mathbb{P}\left[\mathcal{M}( \mathcal{D}^{\prime}) \in \Oc\right]+\delta_c,
\end{align}
\end{Def}
where $\epsilon_c$ accounts for the privacy protection level for the global model, and $\delta_c$ can be viewed as the probability that breaks $(\epsilon_{c},0)$-CDP.

\subsection{Privacy Amplification via Data Subsampling}
Privacy amplification by subsampling~\cite{balle2018privacy} is a key technique for strengthening privacy guarantees without increasing the noise variance. In FL systems, this is achieved by computing model updates over a random subset of the local data rather than the entire dataset.
\begin{Theorem} \label{thm: privacy amplicfication via sampling}
(Privacy amplification by data subsampling~\cite{li2024diffprivate}) Suppose that a random mechanism $\mathcal{M}_i$ for client $i$ in FL is $(\epsilon_\ell, \delta_\ell)$-LDP over local dataset $\mathcal{D}_i$ with size $|\mathcal{D}_i|$. Consider a mechanism $\mathcal{M}_i$ outputs a sample over all subsets $\mathcal{D}_{s} \subseteq \mathcal{D}_i$ with size $| \mathcal{D}_{s} | = Qb$. Then, when $\epsilon_\ell \leq 1$, executing the mechanism $\mathcal{M}_i$ on the subset $\mathcal{D}_{s}$ guarantees $(2q_i\epsilon_\ell, q_i\delta_\ell)$-LDP, where $q_i$ is
\vspace{-0.3cm}
\begin{align}\label{eqn: q_1}
q_i = \left\{\begin{array}{ll}   \frac{Q b}{|\mathcal{D}_i|},     & {\text {WOR,}}    \\
1-(1-\frac{1}{|\mathcal{D}_i|})^{Qb},     & {\text {WR,}}   \end{array}\right.
\end{align}
in which `WOR' and `WR' stand for the data subsampling without and with replacement, respectively.
\end{Theorem}
Theorem~\ref{thm: privacy amplicfication via sampling} shows that the privacy of the local mechanism can be amplified when the effective data subsampling ratio is sufficiently small, e.g., $q_i < 0.5$. Motivated by this observation, it is natural to ask whether a similar amplification effect can also arise at the global model level when only a subset of clients participates in each training round. This is particularly relevant in practical FL systems where client participation may be random or only partially controlled by the PS. To explore this possibility, we next introduce a random client check-in scheme and analyze the resulting CDP amplification effect.

\section{MS-PAFL: Model-Splitting Privacy-Amplified Federated Learning}\label{sec:MS-PAFL}
This section introduces our proposed MS-PAFL framework, which is designed to achieve strong privacy guarantees while preserving high model utility. The framework includes two key components: $(1)$ a random client check-in scheme that enables privacy amplification through client sampling, and $(2)$ a model-splitting strategy that confines privacy-preserving noise to a designated public submodel. We first describe the client check-in mechanism and then present the complete MS-PAFL algorithm.

\subsection{The Proposed Random Client Check-in Scheme}
In typical FL systems, the PS deterministically selects a subset of clients in each round. This server-side control, however, limits the potential for privacy amplification, since the exact set of participants is always known to the PS. To overcome this limitation, we propose a random client check-in scheme. In this paradigm, client participation is self-determined, introducing an additional layer of randomness that strengthens privacy. Specifically, each client independently decides whether to ``check-in'' for a training round with a private probability $p_i^t$ that is hidden from the PS.  For the privacy amplification analysis, we consider an aggregate-level observation model, where the PS has access only to the aggregated public-submodel update and the number of received messages $|\Xc^t|$, rather than individual client updates. Such a setting can be supported by standard secure aggregation primitives~\cite{bonawitz2017practical}, but the detailed protocol is beyond the scope of this work. This uncertainty serves as the fundamental source of privacy amplification via client sampling.

Let $\mathcal{X}^{t}$ denote the random set of clients who participate in  communication round $t$, which is defined as
\begin{align}\label{client sampling pro}
\Xc^t \triangleq \{i \in [N] : \zeta_i^t = 1\},
\end{align}
where $\zeta_{i}^{t} \sim \mathrm{Bernoulli}(p_i^t)$ is an indicator random variable denoting the participation status of client $i$. In particular, $\zeta_i^t=1$ indicates that client $i$ participates in round $t$, and $p_i^t$ denotes the check-in probability of client $i$ at round $t$.

Note that $\Xc^t$ is dynamic and not scheduled by the PS at each round. This characteristic makes our setting fundamentally different from conventional FL systems.
If client $i$ participates in training at round $t$, it first downloads the current global model, performs multiple steps of local SGD, and then uploads its updated model to the PS. The special case $p_i^t = 1,\ \forall i,t$ corresponds to full participation, where all clients join every communication round. By contrast, $p_i^t < 1$ introduces randomness into client participation, which serves as the basis for privacy amplification in our framework.

\begin{figure}[t]
\begin{center}
\resizebox{0.9\linewidth}{!}{\hspace{-0cm}\includegraphics{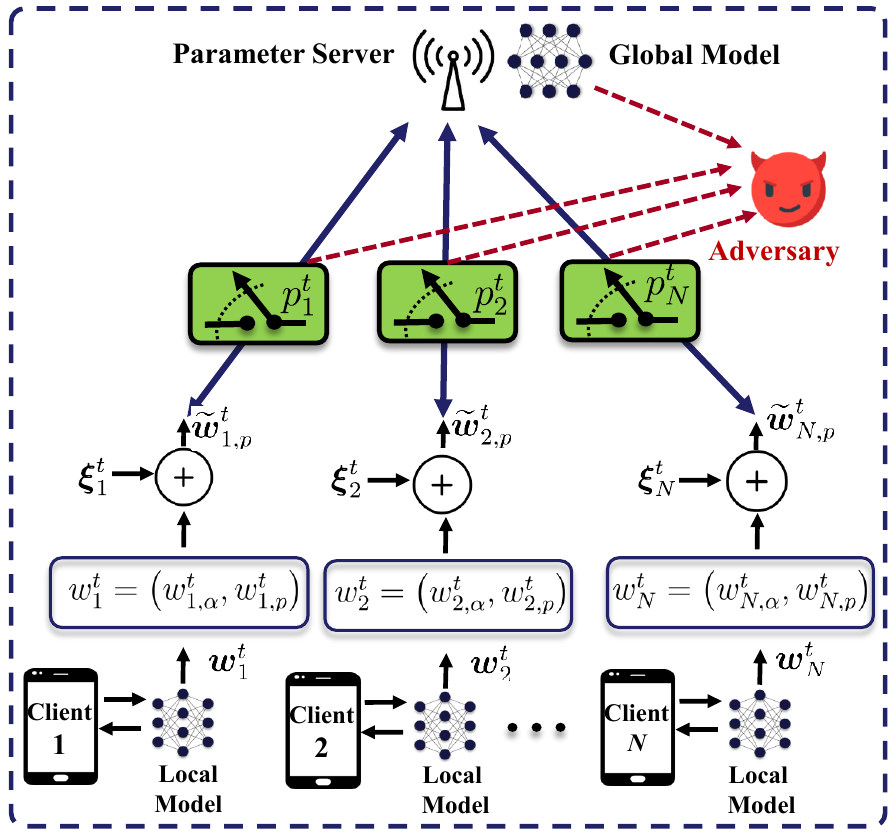}}
\end{center}
\vspace{-0.25cm}
\caption{FL system with random client check-in scheme.}
\label{fig:fedavg_1}
\end{figure}

\subsection{The Proposed Privacy-preserving Scheme: MS-PAFL}
To improve the trade-off between privacy protection and model accuracy,  we propose MS-PAFL, a framework that partitions model parameters into a public submodel for global aggregation and a private submodel retained locally. DP noise is injected only into the public submodel, while the private submodel remains unperturbed to better preserve client-specific information. This selective noising strategy not only improves utility but also provides the basis for our privacy amplification analysis, which leverages both client sampling and data subsampling. By combining structural separation with privacy amplification, MS-PAFL achieves a more favorable privacy--utility trade-off.

The MS-PAFL framework operates in three stages within each communication round: $(i)$ model splitting and local updating, $(ii)$ privacy-preserving upload, and $(iii)$ secure aggregation.

\subsubsection{Model Splitting and Local Updates}
At the beginning of round $t$, each client’s local model $\wb_i^t \in \mathbb{R}^d$ is partitioned into two submodels: a public submodel $\wb_{i,p}^{t} \in \mathbb{R}^{d_p}$ and a private submodel $\wb_{i,\alpha}^{t} \in \mathbb{R}^{d_\alpha}$, where $d_p+d_\alpha=d$.
The public submodel is used for client-server communication and model aggregation, while the private submodel is retained locally to capture client-specific patterns. For analytical tractability, the partition sizes $d_p$ and $d_\alpha$ are assumed to be fixed and commonly known during the training process.
Formally, the client model is denoted by $\wb_i^t = \left(\wb_{i,\alpha}^t, \wb_{i,p}^t\right)$, where $\wb_{i,p}^t$ denotes the visible submodel with dimension $|\wb_{i,p}^t| = d_p$, and $\wb_{i,\alpha}^t$ denotes the invisible submodel with dimension $|\wb_{i,\alpha}^t| = d_\alpha$.

Before injecting calibrated DP noise, gradient clipping is applied to the locally updated public submodel parameters $\wb_{i,p}^t$ to ensure bounded sensitivity. Specifically, if $\|\wb_{i,p}^t\|$ exceeds the threshold, the vector is scaled to norm $\mathcal{C}$; otherwise, it remains unchanged. The model clipping is a standard step in conventional DP-FedAvg, which is essential for sensitivity control in DP-FL.
\begin{align}\label{eqn:clipping}
\hat{{\wb}}_{i , p}^{t}  =  \wb_{i,p}^t/ \max \Big(1, \frac{\big\|\wb_{i,p}^t\big\|}{\mathcal{C}}\Big),
\end{align}
where $\mathcal{C}$ denotes the clipping threshold.

\subsubsection{Privacy-Preserving Upload}
After completing local updates, client $i$ prepares its public submodel for transmission. To guarantee privacy, the calibrated Gaussian noise $\boldsymbol{\xi}_i^{t}$ is added:
\begin{align}\label{eqn:x_0_update_a}
\widetilde{{\wb}}_{i , p}^{t}  =  \hat{{\wb}}_{i, p}^{t} + \boldsymbol{\xi}_i^{t}, \ \boldsymbol{\xi}_{i}^{t} \sim \mathcal{N}\big(\bm{0}, \sigma_{i,{t}}^{2} \mathbf{I}_{d_p}\big),
\end{align}
where $\sigma_{i,t}^{2}$ denotes the noise variance required to guarantee $(\epsilon_{\ell}, \delta_{\ell})$-LDP for the public submodel, as derived in Lemma~\ref{Lemma: global sensitivity}.

Because the private submodel $\wb_{i,\alpha}^t$ is never transmitted to the PS, the privacy cost in MS-PAFL is incurred only through the released public submodel. This selective design enables rigorous privacy protection for the communicated component, while preserving the retained private component from direct perturbation and thereby alleviating unnecessary performance degradation.
At the same time, the private submodel is not claimed to be absolutely secure against all possible inference attacks, since indirect leakage through repeated interactions or correlations with the released public submodel cannot be completely excluded.
\subsubsection{Aggregation at the PS}
The PS aggregates the noisy public submodels $\widetilde{{\wb}}_{i , p}^{t}$ received from participating clients to form the global public model for the next round, denoted by $\overline{\wb}_{p}^{t+1}$.
The resulting aggregation rule is given by
\begin{align}\label{eq:postProcessingKnown}
    \overline{\wb}_{p}^{t+1} & = \left\{\begin{array}{ll}
\frac{1}{|\Xc^t|}  \sum_{i \in \Xc^{t}}  \widetilde{{\wb}}_{i , p}^{t}, & |\Xc^t|>0,\\
\overline{\wb}_{p}^{t}, & |\Xc^t|=0,
\end{array}\right.
\end{align}
where the case  $|\Xc^t|=0$ corresponds to an ``empty round'' under random check-ins.
Under uniform client weights, the above aggregation reduces to standard FedAvg over the randomly participating subset.

\subsubsection{Local Model Recombination}
After aggregation, each client $i$ downloads the updated global public model $\overline{\wb}_{p}^{t+1}$ from the PS and recombines it with the locally retained private submodel to form the local model for the next round:
\begin{align}\label{recombination}
\mathbf{w}_{i}^{t+1,0} = \left(\wb_{i, \alpha}^t,  \overline{\wb}_{p}^{t+1}\right).
\end{align}

This iterative process of local training, selective perturbation, aggregation, and recombination continues until convergence.
A key advantage of MS-PAFL is that the privacy budget is consumed exclusively by the public submodel $\mathbf{w}_p$,  which is the only communicated component and hence the only direct source of information leakage. By confining noise to this component, the framework reduces unnecessary distortion while retaining the privacy benefits brought by client sampling and data subsampling, thereby preserving model utility. The proposed MS-PAFL algorithm is summarized in Algorithm~\ref{alg:scheme}.

\begin{algorithm}[t!]
\caption{Proposed MS-PAFL}
\begin{algorithmic}[1]\label{alg:scheme}
\STATE  \textbf{Input:}  system parameters $b$, $Q$, $N$, $T$.
\STATE Initialize $ \overline\wb^{0} , \{\wb_{i}^{0,1}\}_{i \in[N]}$ and $\Xc^{0}= \emptyset$.
\FOR{$t=0, \ldots, T-1$ }
\STATE $\textbf{Client side:}$
\FOR {$i=1,...,N$}
\STATE Sample $\zeta_i^{t} $ by \eqref{client sampling pro} (client sampling).
\IF{$\zeta_{i}^t=1$}
\FOR{$r= 1, \ldots, Q$}
\STATE  Sample $\mathcal{B}_{i}^{t,r}$ from $\Dc_i$ (data subsampling).
\STATE  Update $\wb_{i}^{t,r}$ by \eqref{eqn:local_Q_update}.
\ENDFOR
\STATE Perform model splitting by $\wb_i^t= \left(\wb_{i, \alpha}^t, \wb_{i, p}^t\right)$.
\STATE Perform model clipping via \eqref{eqn:clipping} to obtain $\hat{{\wb}}_{i, p}^{t}$.
\STATE Update $\widetilde{{\wb}}_{i, p}^{t}$ by \eqref{eqn:x_0_update_a}.
\STATE Send $\widetilde{{\wb}}_{i,p}^{t}$ to the PS for model aggregation.
\ENDIF
\ENDFOR
\STATE $\textbf{PS side:}$
\STATE PS updates global public submodel according to \eqref{eq:postProcessingKnown}.
\STATE Broadcast $\overline{\wb}_{p}^{t+1}$ to all clients.
\ENDFOR
\end{algorithmic}
\end{algorithm}

\section{Privacy Analysis}\label{subsec:privacy_analysis}
In this section, we present a rigorous privacy analysis of the proposed MS-PAFL framework. Our goal is to quantify the privacy guarantees for both the local and global models, explicitly considering the two sources of privacy amplification:  random client check-in and local data subsampling. The analysis proceeds in two steps. First, we establish the single-round privacy guarantee for the released global public submodel, which is the only channel of potential leakage under model splitting. We then extend this result to derive the total privacy loss for both the local and global models.

\subsection{Single-Round Privacy Guarantee}
We begin by analyzing the privacy guarantee in a single communication round. Each client injects Gaussian noise into its public submodel update, ensuring that the local mechanism satisfies $(\epsilon_\ell,\delta_\ell)$-LDP. We then examine how this guarantee propagates through aggregation at the PS, leading to the overall privacy protection of the global public model. The result is summarized in the following theorem.
\begin{Theorem}\label{Thm: cdp for global model}
Suppose that the local update of each client $i$ satisfies $(\epsilon_\ell, \delta_\ell)$-LDP. In MS-PAFL, each client $i$ participates independently with probability $p_i^t$ (cf. \eqref{client sampling pro}) and applies a data subsampling ratio $q_i$ (cf. \eqref{eqn: q_1}). Then, the aggregated global public submodel $\overline{\wb}_{p}^{t+1}$ in round $t$ satisfies $(\epsilon_{c}^t, \delta_{c}^t)$-CDP, where
\begin{subequations} \label{eqn:thm_parameters}
\begin{align}
\epsilon_{c}^t \leq \,&\, \max_{i} \Big\{\ln \Big[1+\frac{p_i^t}{1-\delta^{\prime}}\big(\exp(2 q_i  \epsilon_\ell)-1\big)\Big] \Big\},   \label{eqn:10a} \\
\delta_{c}^t =\,&\,\max_{i } \Big\{\delta^{\prime}+\frac{ p_i^t q_i   \delta_\ell}{1-\delta^{\prime}} \Big\},  \label{eqn:10b}
\end{align}
\end{subequations}
with
\begin{align}
\delta^{\prime} \triangleq \mathbb{P}(|\,|\Xc^t|-\mu_{_{\Xc}}^t| \geq \beta N) \leq 2 \exp(-2 \beta^{2} N),  \label{Lemma: Hoeffding’s Inequality}
\end{align}
where $\mu_{_{\Xc}}^t=\sum_{i=1}^{N} p_{i}^{t}$ is the expected number of participating clients at round $t$, and $\beta > 0$ is a deviation parameter (typically chosen such that $0<\beta<1$).
\end{Theorem}

\vspace{-0.25cm}
\noindent \textit{Proof:} The proof is inspired by the work in \cite{mohamed2021privacy}. However, that analysis does not address the coupled setting introduced by model splitting and privacy amplification.  For completeness, we restate the proof in Appendix~\ref{sec:Proof of the theorem3}. $\hfill\blacksquare$

Theorem~\ref{Thm: cdp for global model} provides the core of our privacy analysis, showing that the joint use of client sampling and data subsampling yields substantial privacy amplification. The resulting bound on $\epsilon_{c}^t$ is provably tighter than that achievable by either sampling scheme alone, thereby reducing privacy leakage under the same level of injected noise.

For the scenario where all clients adopt identical participation probabilities and data subsampling ratios, the privacy guarantee admits a more concise form.
\begin{Corollary}\label{cor::uniform_sampling}
(Uniform sampling) If $p_i^t = p$ and $q_i = q$ for all clients, and each public update satisfies $(\epsilon_\ell, \delta_\ell)$-LDP, then the global model at round $t$ satisfies $(\epsilon_{c}^t, \delta_{c}^t)$-CDP, where
\begin{subequations}\label{eqn:thm_parameters_uniform}
\begin{align}
\epsilon_{c}^t \leq \,&\, \ln \Big[1+\frac{p}{1-\delta^{\prime}}\big(\exp(2 q  \epsilon_\ell)-1\big)\Big],   \label{eqn:10a_c1a} \\
\delta_{c}^t =\,&\,  \delta^{\prime}+\frac{ p q   \delta_\ell}{1-\delta^{\prime}},  \label{eqn:10b_c2b}
\end{align}
\end{subequations}
with $\delta'$ defined in \eqref{Lemma: Hoeffding’s Inequality}.
\end{Corollary}

For \eqref{eqn:10a_c1a}, the CDP of released global model can be upper bounded as
\begin{align}
\epsilon_{c}^t  \leq  \frac{p}{1-\delta^{\prime}}\big(\exp(2 q  \epsilon_\ell)-1\big) \overset{(a)}{\leq}  \frac{2 q  \epsilon_\ell p}{1-\delta^{\prime}}  = \mathcal{O}\big(p q \epsilon_\ell\big), \label{eqn:beta}
\end{align}
where $(a)$ follows from the first-order approximation $e^x-1 \thickapprox x$ for small $x$.

\begin{Remark}\label{rmk:split_impact}
(Impact of model splitting) Since DP noise is injected only into the public submodel, while the private submodel is retained locally without perturbation, the overall perturbation imposed on the complete model is reduced compared with conventional full-model DP-FL schemes. As a result, the proposed MS-PAFL is more favorable for preserving training stability and mitigating the loss of convergence performance, thereby providing a structural advantage in balancing privacy protection and model utility.
\end{Remark}

\begin{Remark}\label{rmk:Remark1}
(Privacy and convergence trade-off) Theorem~\ref{Thm: cdp for global model} reveals a fundamental trade-off: smaller participation probabilities $p_i^t$ and data subsampling ratios $q_i$ lead to a tighter bound on $\epsilon_c^t$, thus improving privacy, but they also reduce the effective training data per round and slow convergence.
\end{Remark}

\begin{Remark}
(Computational analysis of MS-PAFL) The proposed MS-PAFL adopts a fixed and commonly known partition rule once the splitting ratio has been specified, rather than performing per-round dynamic repartitioning. Hence, the additional computation for extraction, perturbation, and recombination scales as $O(d_p)$ per round, while the communication cost is reduced from $O(d)$ to $O(d_p)$.
\end{Remark}

\begin{Remark}\label{rmk:joint_amplification}
(Advantage of joint sampling amplification) The~\eqref{eqn:beta} shows that, under the proposed framework, the single-round central privacy loss scales as $\mathcal{O}(p q \epsilon_\ell)$. This is substantially tighter than the $\mathcal{O}(\epsilon_\ell)$ scaling reported in~\cite{mohamed2021privacy,balle2020privacy} without jointly exploiting client sampling and data subsampling.
\end{Remark}

\subsection{Total Privacy Loss Analysis}\label{sec:Total privacy leakage}
While the previous analysis characterizes the privacy loss for a single round, it is crucial to quantify the total privacy cost across the entire training process. To this end, we employ the moments accountant method~\cite{abadi2016deep} to track the total privacy loss for local clients and the strong composition theorem to bound the total privacy loss of the global model.

\subsubsection{Total Privacy Loss for Local Clients}

\begin{Theorem}\label{Thm:total_budget}
Suppose that each client $i, \forall i \in [N]$, in Algorithm~\ref{alg:scheme} participates independently with a uniform probability $p_i$, and that each communication round guarantees $(\epsilon_{\ell}, \delta_{\ell})$-LDP. Then, the total privacy loss $\bar{\epsilon}_{i}^T$ of client $i$ after $T$ communication rounds satisfies
\begin{small}
\begin{align} \label{eqn:total_privacyloss}
\bar{\epsilon}_{i}^T = c_{0} q_i \epsilon_{\ell} \sqrt{\frac{p_i T}{1-q_i} }, \forall i \in [N],
\end{align}
\end{small}
\vspace{-0.1cm}
where $ c_{0} >0$ is a constant and $q_i$ is given in \eqref{eqn: q_1}.
\end{Theorem}
\textit{Proof:} The proof mainly follows the approach in \cite{li2025differentially}, but that work did not consider random client participation. For completeness, we briefly restate the proof in Appendix~\ref{subsec:the_Proof_of_Theorem_dp}. $\hfill\blacksquare$

Theorem~\ref{Thm:total_budget} offers a local-mechanism perspective on privacy. Equation~\eqref{eqn:total_privacyloss} shows that a client’s total privacy loss ($\bar{\epsilon}_i^T$) is directly tied to its participation frequency, scaling as $\mathcal{O}(q_i\sqrt{p_i T})$. This is substantially tighter than the $\mathcal{O}(\sqrt{T})$ bound obtained without privacy amplification~\cite{abadi2016deep,li2020secure}.

\begin{Remark}\label{Remark qi}
(Impact of data subsampling strategies) According to Theorem~\ref{Thm: cdp for global model}, data subsampling with replacement provides stronger privacy amplification than sampling without replacement due to $1-(1- {1}/{|\mathcal{D}_i|})^{Qb} \leq {(Q b)}/{|\mathcal{D}_i|}$, which yields a smaller effective sampling ratio $q_i$ under with replacement. A smaller $q_i$ directly tightens the privacy bound, thereby enhancing the protection level for a fixed privacy budget. Consequently, data subsampling with replacement generally achieves  a slightly better privacy--utility trade-off than without replacement, especially when $q_i$ is small.
\end{Remark}

\subsubsection{Total Privacy Loss for Global Model}
By applying the strong composition theorem~\cite{5670947}, the total privacy loss of the global model, denoted by $\overline{\epsilon}_{c}^T$, can be bounded as follows.
\begin{Theorem}\label{Thm:composition}
Suppose that the global model $\overline{\wb}^{t}$ guarantees $(\epsilon_{c}^t, \delta_{c}^t)$-DP at each round $t$. Then, after $T$ communication rounds, the sequence of models $\overline{\wb}^{t}$ achieves ($\overline{\epsilon}_{c}^T, T\delta_{c}^{\prime} + \widetilde\delta$)-DP, where
\begin{subequations}
\begin{align}
&\overline{\epsilon}_{c}^T = \sqrt{2T\ln(1/\widetilde\delta)}\epsilon_{c}^{\prime} + T \epsilon_{c}^{\prime} (\exp(\epsilon_{c}^{\prime})-1),  \label{eqn:Thm4_a}\\
&\widetilde\delta \in (0,1], \ \epsilon_{c}^{\prime} =\max_{t} \ \epsilon_{c}^t,\ \delta_{c}^{\prime} =\max_{t} \ \delta_{c}^t.  \label{eqn:Thm4_b}
\end{align}
\end{subequations}
\end{Theorem}
\textit{Proof:} This result follows directly from the strong composition theorem in~\cite{5670947}, and is therefore omitted here for brevity.

Theorem~\ref{Thm:composition} provides the essential tool to bound the total privacy loss for the global model.
The key insight of the strong composition theorem is that the total privacy loss grows \emph{sublinearly} with $T$, which is highly desirable in FL systems.

\begin{Remark}
The total privacy loss for the global model $\overline{\epsilon}_{c}^T$ depends on the worst-case single-round privacy loss $\epsilon_{c}^{\prime}$ (cf. \eqref{eqn:Thm4_b}). By structurally splitting models, injecting noise only into the public submodel, and leveraging both client and data subsampling, MS-PAFL  can effectively reduce this worst-case leakage. As a result, for the same target accuracy, it achieves a lower total privacy cost compared with conventional FL, thereby yielding a more favorable privacy--utility trade-off.
\end{Remark}

\section{Simulation Results}\label{sec:Experimental Details}
In this section, we conduct extensive experiments to validate the proposed analysis and support our theoretical results. Specifically, the experiments are designed to: $(i)$ demonstrate the privacy amplification effect brought by client sampling and data subsampling, $(ii)$ quantify the influence of different sampling parameters on both per-round and cumulative privacy loss, and $(iii)$ empirically examine the role of structural model splitting under different splitting ratios $d_p/d$.

\subsection{Experimental Setup}
{\bf Datasets:}  We evaluate the proposed method on the Adult~\cite{blake1998uci} and MNIST~\cite{website_MNIST} datasets. The MNIST dataset contains 60,000 training samples and 10,000 testing samples, while the Adult dataset contains 32,561 training samples and 16,281 testing samples. For the Adult dataset, missing entries are imputed using the most frequent value of the corresponding feature, and all features are normalized before training. In all experiments, the training data are distributed over $N=100$ clients.

Unless otherwise stated, all experiments are conducted under non-iid data distribution. For MNIST, we adopt the non-iid partition strategy in~\cite{li2019convergence}, where each client is assigned samples from only a limited number of classes, thereby yielding heterogeneous local label distributions. The degree of heterogeneity is controlled by the number of class labels available at each client. For Adult, following~\cite{li2020secure}, the training samples are distributed across the $N=100$ clients such that each client contains samples from only one class.
\\
{\bf Parameter setting:} For all experiments, we set the target privacy parameter $\delta = 10^{-4}$. The Hoeffding's inequality parameter is set to $\beta = 0.25$, yielding the bound $\delta^{\prime} = 2 \exp(-2 \beta^2 N)$ (cf. \eqref{Lemma: Hoeffding’s Inequality}). To guarantee $(\epsilon_\ell, \delta_\ell)$-LDP for each client's update, Gaussian noise $\xib_{i}^{t}$ with variance $\sigma_{i,t}^2$ is added to the public submodel. All clients are assumed to have the same check-in probability, $p_i = p_i^t, \forall t \in [T]$. The total privacy of the global model, $\overline{\epsilon}_{c}^T$, is estimated using the strong composition mechanism from Theorem \ref{Thm:composition}. For model splitting, we adopt a fixed and commonly known partition rule during training, where each local model is divided into a private submodel and a public submodel according to a pre-specified splitting ratio $d_\alpha:d_p$.  Following the rule of gradient clipping in \cite{abadi2016deep},  we select the value of $\mathcal{C}$ by taking the median of the norms of the unclipped gradients over the training process.  The CNN-based experiment follows the same privacy setting, while its network architecture and training parameters are specified together with the corresponding results.

\begin{figure*}
	\begin{minipage}[b]{0.48\linewidth}
		\includegraphics[scale=0.50]{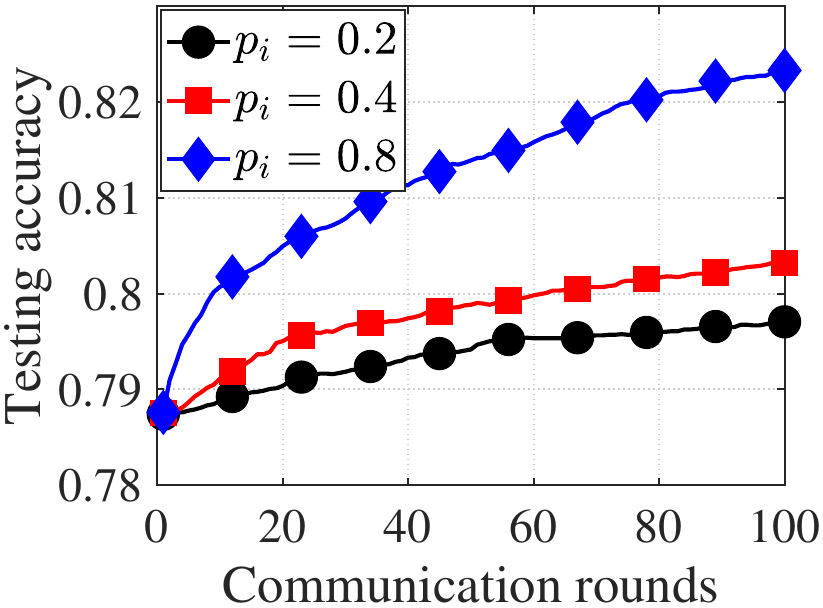}
		\centerline{{(a)  Adult, WR}}\medskip
	\end{minipage}
	\hfill
	\begin{minipage}[b]{0.48\linewidth}
		\includegraphics[scale=0.50]{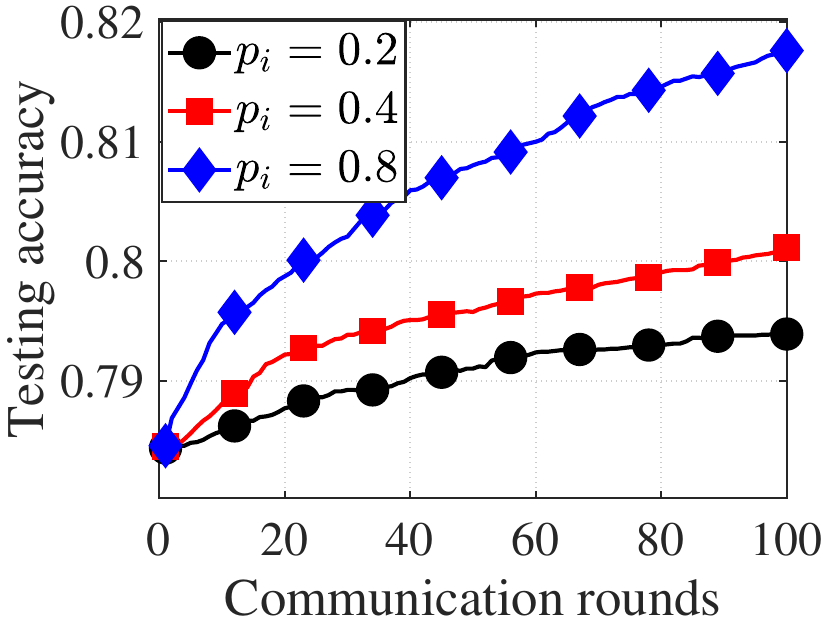}
		\centerline{{(b)  Adult, WOR}}\medskip
	\end{minipage}
    \hfill
    \begin{minipage}[b]{0.48\linewidth}
		\includegraphics[scale=0.50]{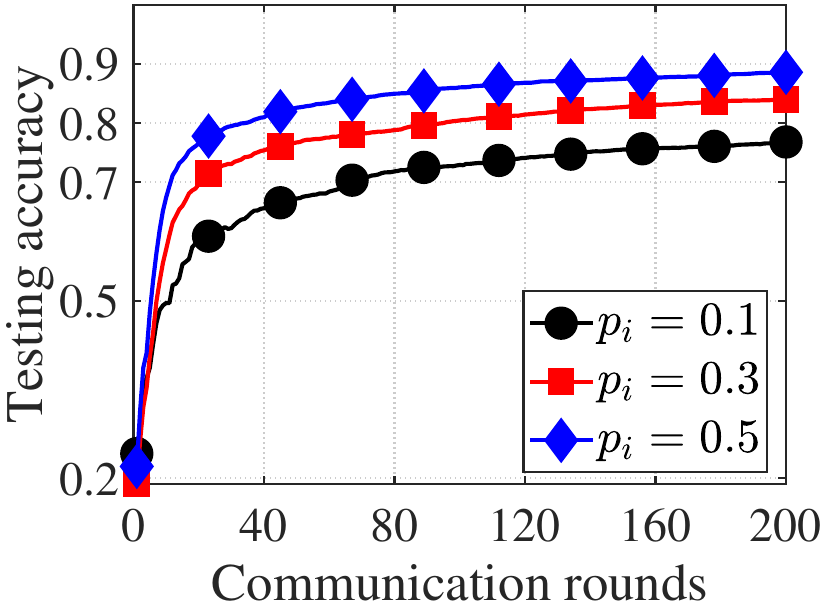}
		\centerline{{(c)  MNIST, WR}}\medskip
	\end{minipage}
	\hfill
	\begin{minipage}[b]{0.48\linewidth}
		\includegraphics[scale=0.50]{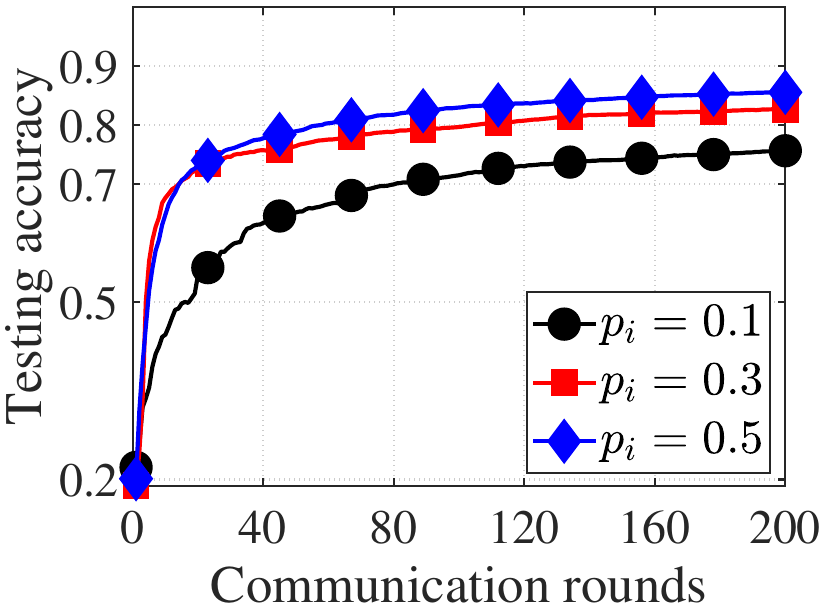}
		\centerline{{(d) MNIST, WOR}}\medskip
	\end{minipage}
	\caption{Testing accuracy versus communication rounds of the proposed MS-PAFL under the WR and WOR cases. Subfigures (a)–(b) and (c)–(d) correspond to the Adult and MNIST datasets, respectively, under different client participation levels. }
	\vspace{-0.05cm}
	\label{Fig:client_checkin}
\end{figure*}

\noindent $\textbf{Benchmark algorithms:}$
We compare the proposed MS-PAFL framework with DP-FedAvg~\cite{li2020secure}, which does not employ model splitting or privacy amplification. In addition, to better isolate the contribution of structural partitioning, we report ablation results of MS-PAFL under different splitting ratios $d_\alpha/d_p$, so as to examine how the choice of public/private decomposition affects the resulting privacy--utility trade-off.

\subsection{Impact of the Random Client Participation $p_i$ }
Figure~\ref{Fig:client_checkin} shows the testing accuracy of the proposed algorithm under different client participation probabilities $p_i$ on both the Adult and MNIST datasets. In these experiments, the number of local update steps is set to $Q=5$, and the model splitting ratio is fixed as $d_\alpha:d_p=1:9$. The cumulative privacy budgets are set to $\overline{\epsilon}_c^T=2$ for the Adult dataset and $\overline{\epsilon}_c^T=20$ for the MNIST dataset.  Two main observations can be made from Fig.~\ref{Fig:client_checkin}.
First, under both the WR and WOR cases, the testing accuracy consistently improves as $p_i$ increases. This trend is observed on both datasets and is particularly evident for MNIST, where a higher participation level not only accelerates convergence but also leads to a higher final accuracy. This observation is consistent with Remark~\ref{rmk:Remark1}.

Second, the comparison between the WR and WOR cases shows that the local data subsampling strategy also affects the convergence behavior, although its impact is less pronounced than that of client participation. The WR case achieves slightly better performance than the WOR case in several settings, especially in terms of the final testing accuracy. Although the improvement is moderate, it remains consistent with our theoretical analysis, which suggests that data subsampling with replacement may yield a more favorable privacy--utility trade-off than sampling without replacement. Overall, these results highlight the need to jointly balance participation probability, privacy protection level, and local sampling design in practical FL systems.

\begin{figure}
\begin{minipage}[b]{0.48\linewidth}
\includegraphics[scale=0.50]{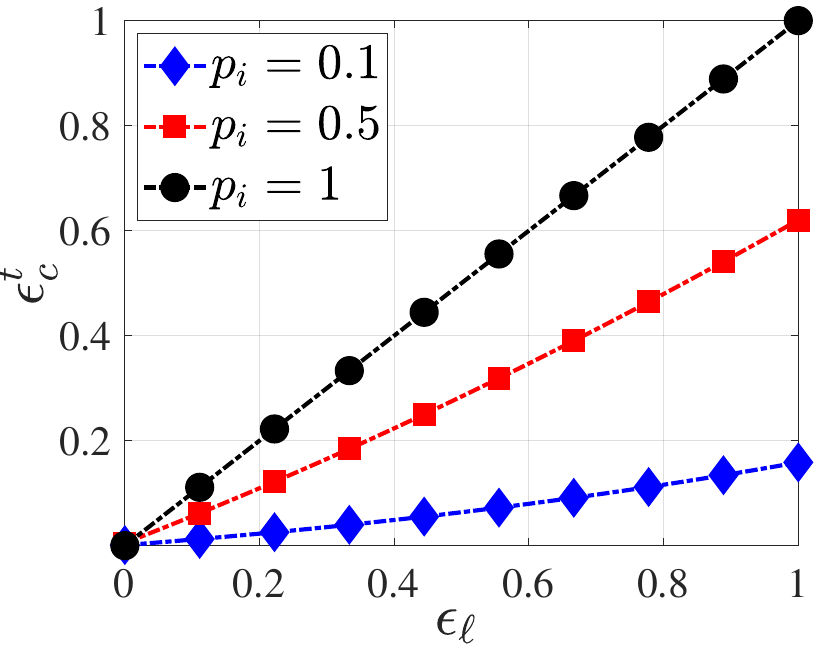}
\centerline{{(a)}}\medskip
\vspace{-0.35cm}
\end{minipage}
\hspace{0.1cm}
\begin{minipage}[b]{0.48\linewidth}
\includegraphics[scale=0.50]{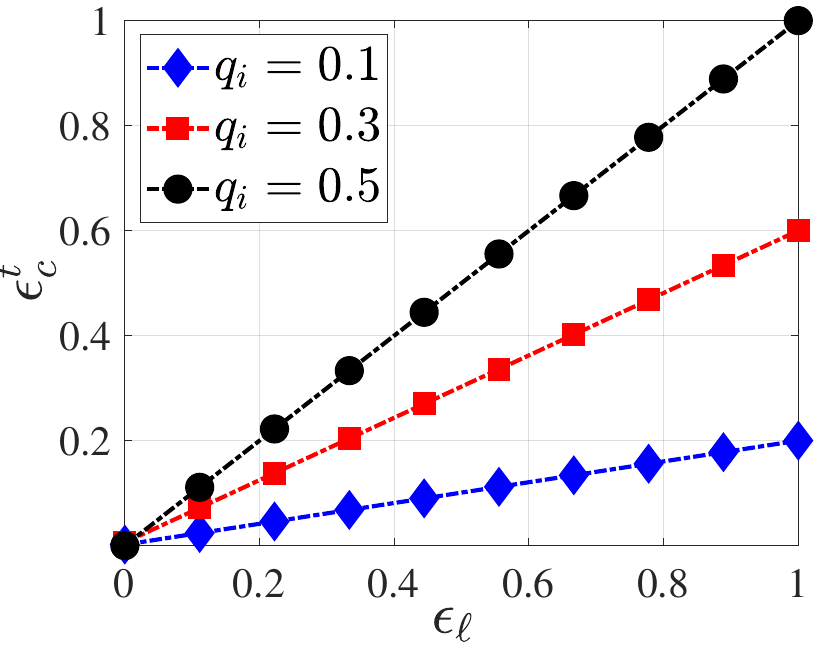}
\centerline{{(b)}}\medskip
\vspace{-0.35cm}
\end{minipage}
\caption{Single-round central privacy loss $\epsilon_{c}^t$ versus the local privacy level $\epsilon_{\ell} \in [0,1]$: (a) different client participation probabilities $p_i \in \{0.1,0.5,1\}$; (b) different local data subsampling ratios $q_i \in \{0.1,0.3,0.5\}$.}
\vspace{-0.05cm}
\label{Fig:Amplification_for_global_model}
\end{figure}

\subsection{Validation of Privacy Amplification Mechanisms}
Figure~\ref{Fig:Amplification_for_global_model} illustrates the single-round central privacy loss $\epsilon_c^t$ as a function of the local privacy level $\epsilon_\ell$ under model splitting ratio $d_\alpha:d_p=1:9$.  Specifically, Fig.~\ref{Fig:Amplification_for_global_model}(a) varies the client participation probability over $p_i \in \{0.1,0.5,1\}$ to examine the effect of client sampling, whereas Fig.~\ref{Fig:Amplification_for_global_model}(b) varies the local data subsampling ratio over $q_i \in \{0.1,0.3,0.5\}$ to evaluate the impact of data sampling. Two consistent trends can be observed: $(i)$ decreasing the client participation probability $p_i$ significantly reduces $\epsilon_c^t$, and $(ii)$ decreasing the local data subsampling ratio $q_i$ similarly leads to a tighter privacy bound. These results directly corroborate Theorem~\ref{Thm: cdp for global model}, which shows that the combination of client sampling and data subsampling yields a tighter privacy bound than either mechanism alone. In particular, the case $p_i=1$ serves as the baseline and results in the largest central privacy loss, whereas smaller values of $p_i$ or $q_i$ increase the uncertainty regarding which clients or local samples contribute to the update, thereby strengthening the privacy guarantee under the same local privacy level.

\begin{figure}
\begin{minipage}[b]{0.48\linewidth}
\includegraphics[scale=0.50]{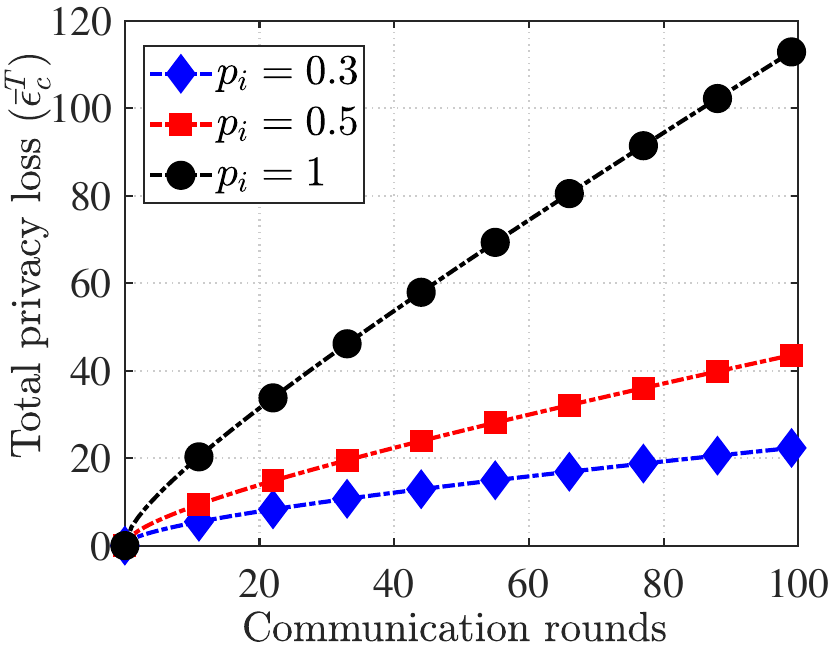}
\centerline{{(a)}}\medskip
\vspace{-0.35cm}
\end{minipage}
\hspace{0.1cm}
\begin{minipage}[b]{0.48\linewidth}
\includegraphics[scale=0.50]{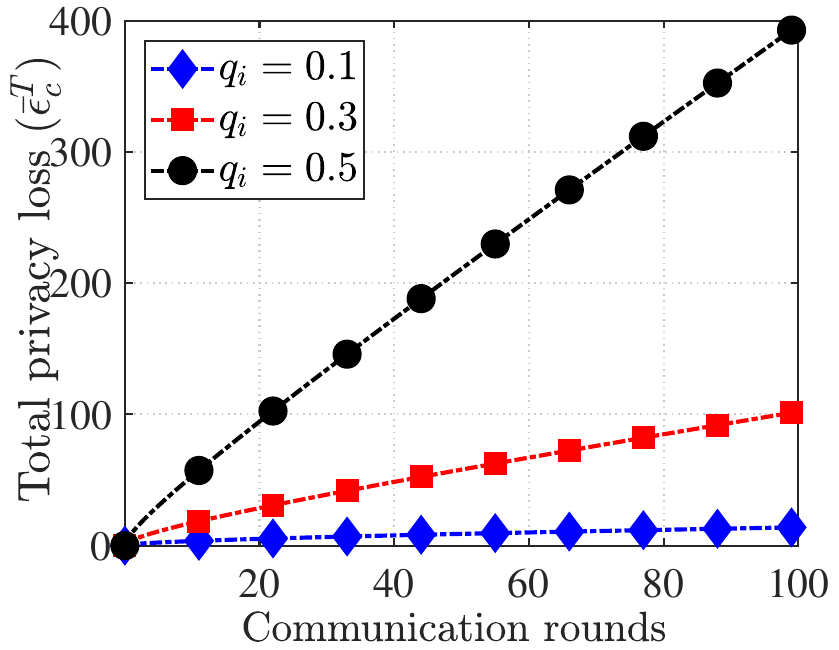}
\centerline{{(b)}}\medskip
\vspace{-0.35cm}
\end{minipage}
\caption{Total privacy loss of the global model $\overline{\epsilon}_{c}^T$ under WOR sampling: (a) different values of $p_i$ with fixed $q_i=0.2$; (b) different values of $q_i$ with fixed $p_i=0.5$.}
\vspace{-0.05cm}
\label{Fig:Amplification_comparison}
\end{figure}

Figure~\ref{Fig:Amplification_comparison} depicts the total privacy loss of the global model, $\overline{\epsilon}_{c}^T$, under the WOR sampling strategy. Specifically, in Fig.~\ref{Fig:Amplification_comparison}(a), we fix $q_i=0.2$ and vary $p_i$ to show the effect of client participation on the cumulative privacy cost, while in Fig.~\ref{Fig:Amplification_comparison}(b), we fix $p_i=0.5$ and vary $q_i$ to examine the impact of local data subsampling. The results confirm that the single-round privacy amplification effect accumulates across multiple communication rounds. Smaller values of $p_i$ or $q_i$ lead to a substantially lower total privacy loss, which is consistent with Theorem~\ref{Thm:total_budget} and Remark~\ref{rmk:joint_amplification}.
From a practical perspective, these results provide three insights. First, $p_i$ and $q_i$ should be jointly selected to satisfy a target privacy budget while maintaining adequate convergence speed. Second, the privacy--utility trade-off is explicit, since reducing $p_i$ and $q_i$ lowers the total privacy loss but may also slow down training. Third, the local sampling strategy is also non-negligible, since sampling with replacement generally provides stronger amplification than sampling without replacement, thereby further reducing the total privacy loss under the same privacy level, which is consistent with Remark~\ref{Remark qi}.

\subsection{Comparison of Total Privacy Loss under Different Data Subsampling Strategies}
\begin{figure}[t]
\begin{minipage}[b]{0.48\linewidth}
\includegraphics[scale=0.50]{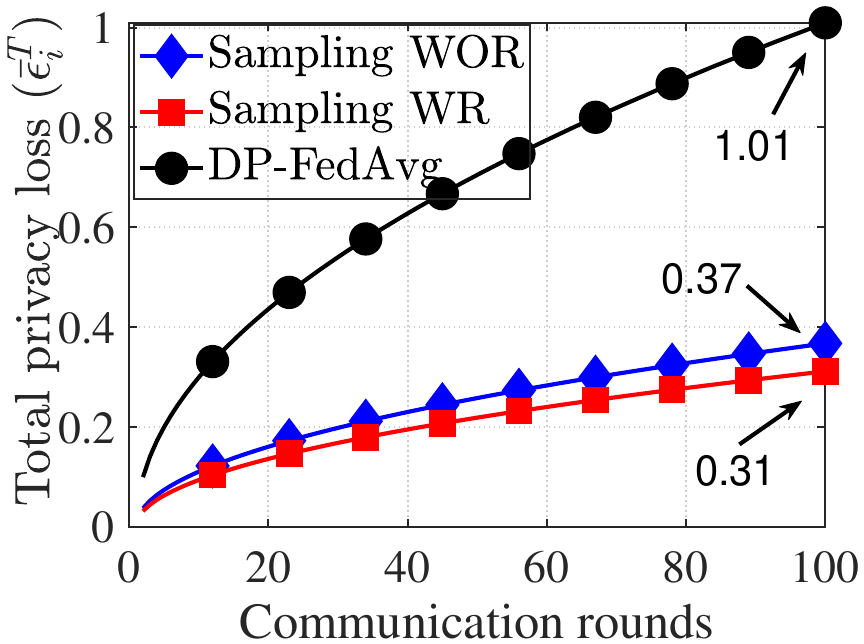}
\centerline{(a) }\medskip
\vspace{-0.35cm}
\end{minipage}
\hspace{0.05cm}
\begin{minipage}[b]{0.48\linewidth}
\includegraphics[scale=0.50]{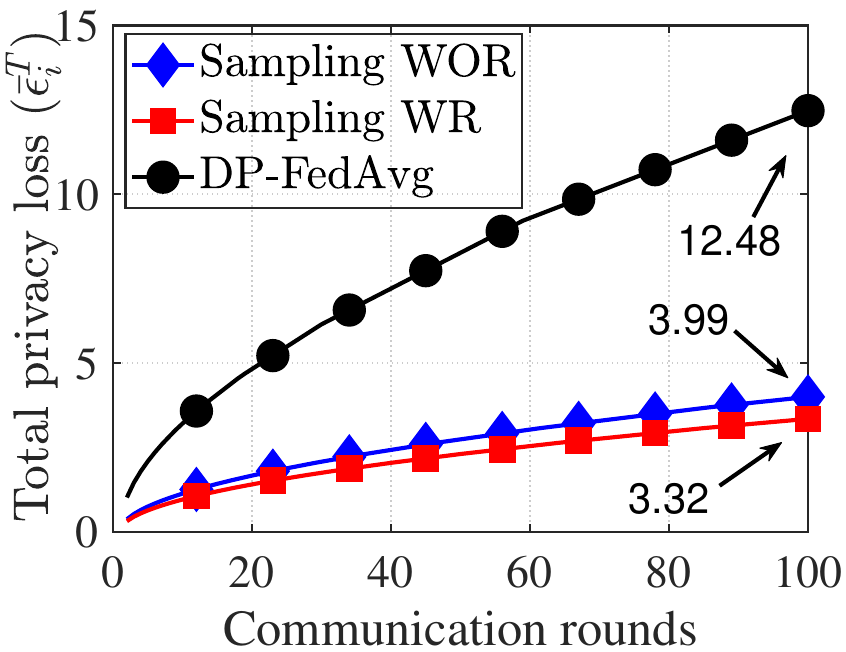}
\centerline{(b)}\medskip
\vspace{-0.35cm}
\end{minipage}
\caption{Total privacy loss of the local client $\overline{\epsilon}_{i}^T$ under different local data subsampling strategies: (a) $\epsilon_{\ell}=0.1$ and (b) $\epsilon_{\ell}=1$.}
\label{Fig:privacyloss_PS}
\end{figure}

\begin{figure}[t]
\begin{minipage}[b]{0.48\linewidth}
\includegraphics[scale=0.50]{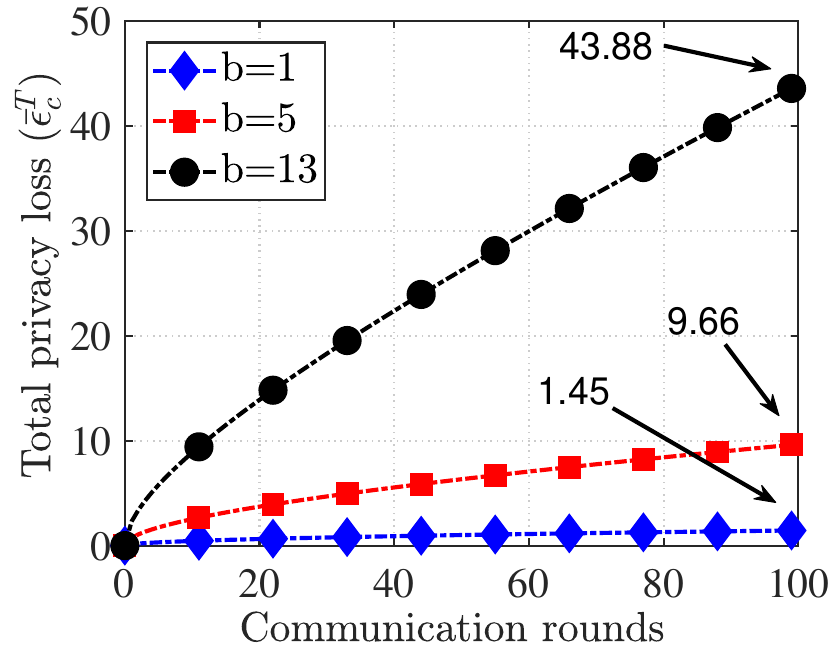}
\centerline{(a) }\medskip
\vspace{-0.35cm}
\end{minipage}
\hspace{0.05cm}
\begin{minipage}[b]{0.48\linewidth}
\includegraphics[scale=0.50]{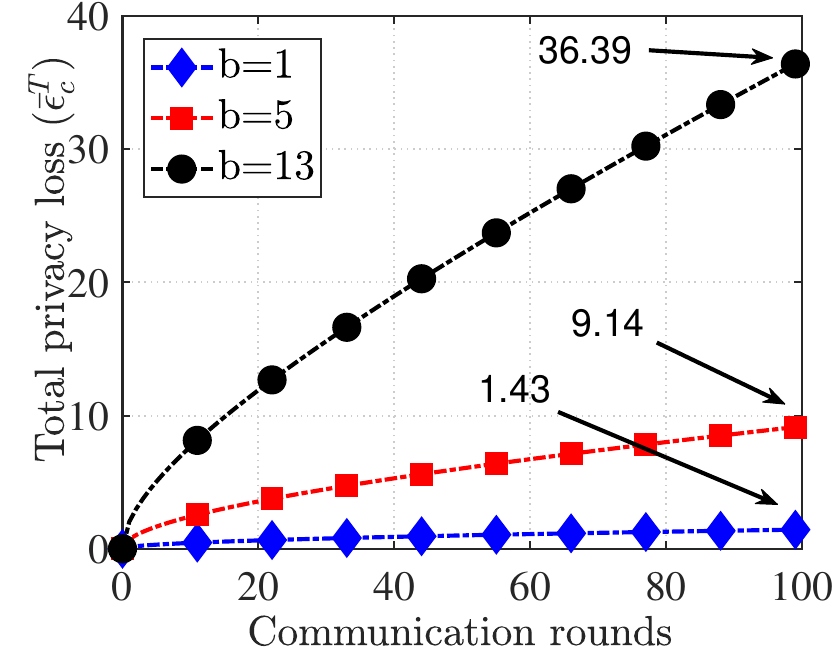}
\centerline{(b)}\medskip
\vspace{-0.35cm}
\end{minipage}
\caption{Total privacy loss of the global model $\overline{\epsilon}_{c}^T$ under different mini-batch sizes $b \in \{1,5,13\}$: (a) WOR and (b) WR.}
\label{Fig:privacyloss_batch}
\end{figure}

Figures~\ref{Fig:privacyloss_PS} and~\ref{Fig:privacyloss_batch} compare the total privacy loss of the local client $\bar{\epsilon}_{i}^T$  and the global model $\bar{\epsilon}_{c}^T$ under  model splitting ratio $d_\alpha:d_p=1:9$. In these experiments, the number of local update steps is fixed at $Q=5$. Figure~\ref{Fig:privacyloss_PS} presents the total privacy loss of the local client for $\epsilon_{\ell}=0.1$ in Fig.~\ref{Fig:privacyloss_PS}(a) and $\epsilon_{\ell}=1$ in Fig.~\ref{Fig:privacyloss_PS}(b). It is observed that sampling with replacement consistently results in a smaller total privacy loss than sampling without replacement, and that both strategies outperform the baseline DP-FedAvg without privacy amplification. Moreover, the advantage of sampling with replacement remains evident as the local privacy level increases, which is consistent with Theorem~\ref{Thm:total_budget} and Remark~\ref{Remark qi}.

Figure~\ref{Fig:privacyloss_batch} shows the total privacy loss of the global model under different mini-batch sizes $b \in \{1,5,13\}$. One can observe from the figure that the total privacy loss increases with $b$, highlighting the important role of local data subsampling in controlling the cumulative privacy cost of the global model. In addition, for the same $b$, sampling with replacement consistently yields a smaller privacy loss than sampling without replacement, further verifying the stronger privacy amplification effect of the WR strategy.

\begin{figure}
	\begin{minipage}[b]{0.48\linewidth}
		\includegraphics[scale=0.50]{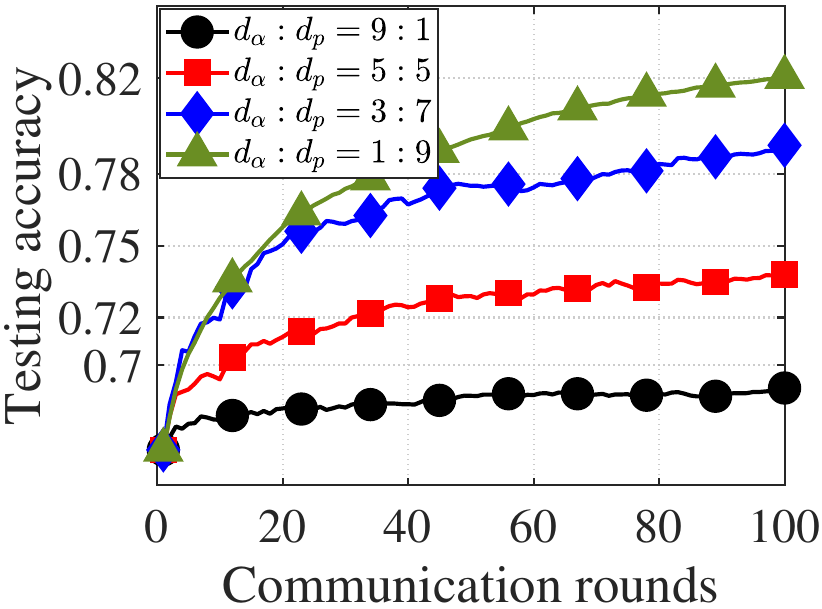}
		\centerline{{(a)  Adult, WR}}\medskip
	\end{minipage}
	\hfill
	\begin{minipage}[b]{0.48\linewidth}
		\includegraphics[scale=0.50]{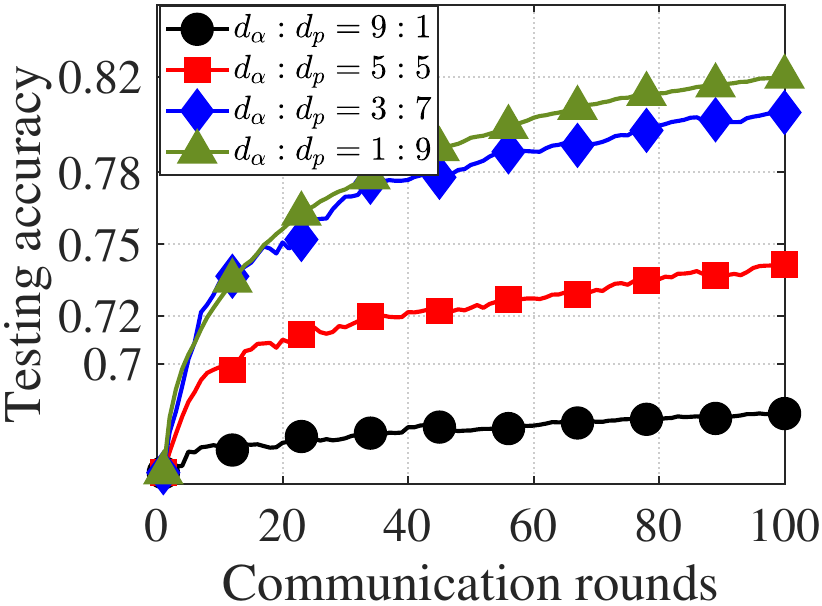}
		\centerline{{(b)  Adult, WOR}}\medskip
	\end{minipage}
    \hfill
    \begin{minipage}[b]{0.48\linewidth}
		\includegraphics[scale=0.50]{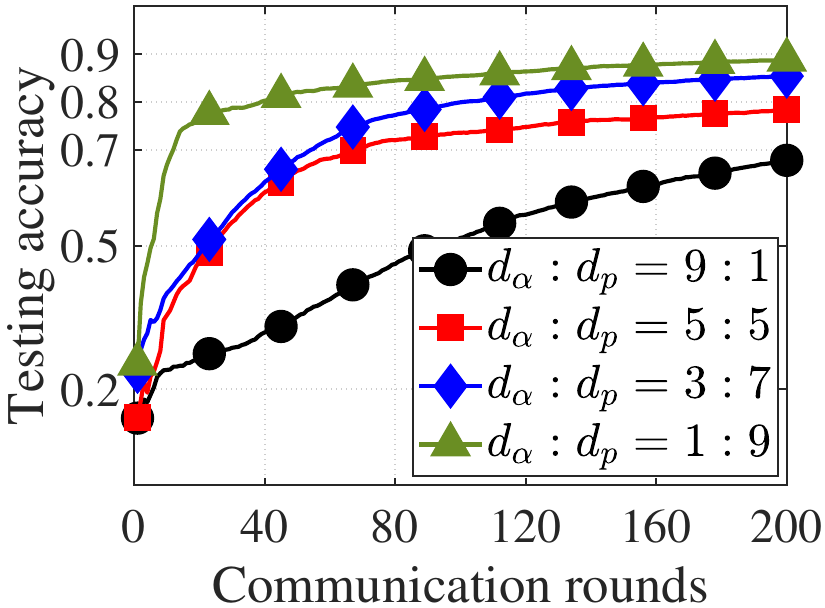}
		\centerline{{(c)  MNIST, WR}}\medskip
	\end{minipage}
	\hfill
	\begin{minipage}[b]{0.48\linewidth}
		\includegraphics[scale=0.50]{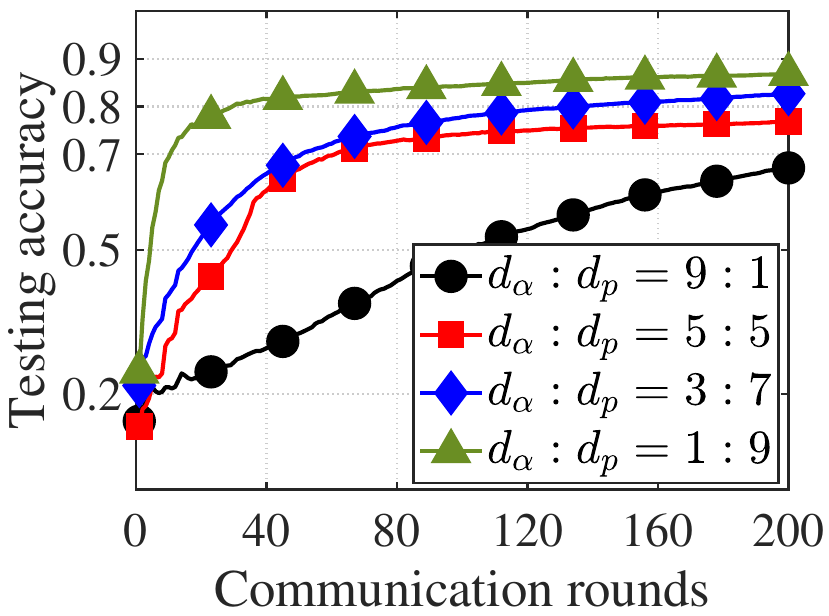}
		\centerline{{(d) MNIST, WOR}}\medskip
	\end{minipage}
	\caption{Testing accuracy versus communication rounds for different model splitting ratios $d_\alpha:d_p$: (a) Adult under WR, (b) Adult under WOR, (c) MNIST under WR, and (d) MNIST under WOR. }
	\vspace{-0.05cm}
	\label{Fig:ablation}
\end{figure}

\subsection{Ablation on Model Splitting Ratio and Strategy}
To further examine the effect of model splitting, we conduct an ablation study under different splitting ratios $d_\alpha:d_p$, where $d_\alpha$ and $d_p$ denote the dimensions of the invisible and visible submodels, respectively. In this experiment, the number of local update steps is fixed at $Q=5$, and the number of participating clients is set to $N=50$. The cumulative privacy budgets are specified as $\overline{\epsilon}_c^T=2$ for the Adult dataset and $\overline{\epsilon}_c^T=20$ for the MNIST dataset. As shown in Fig.~\ref{Fig:ablation}, the testing accuracy on both datasets consistently improves as the splitting ratio varies from $d_\alpha:d_p=9:1$ to $d_\alpha:d_p=1:9$. This result suggests that a larger visible submodel is more favorable for global aggregation, since it allows more informative parameters to be shared and jointly optimized. In contrast, when the invisible submodel dominates, e.g., $d_\alpha:d_p=9:1$, the amount of shared trainable information is significantly reduced, resulting in noticeable performance degradation.

Moreover, a similar performance ordering is observed under both WR and WOR, indicating that the effect of model splitting remains robust with respect to the local sampling strategy. Although the performance gap between WR and WOR is relatively small, the splitting ratio has a much more pronounced impact, particularly on the MNIST dataset, where larger visible submodels lead to faster convergence and higher testing accuracy. Overall, these results further confirm the inherent privacy--utility trade-off in the proposed framework: increasing the proportion of the invisible submodel enhances local information protection, but reduces the effective shared model capacity and consequently weakens collaborative training performance. Therefore, an appropriate splitting ratio is essential for achieving a desirable balance between privacy preservation and learning utility.

\begin{figure}
\begin{minipage}[b]{0.48\linewidth}
\includegraphics[scale=0.50]{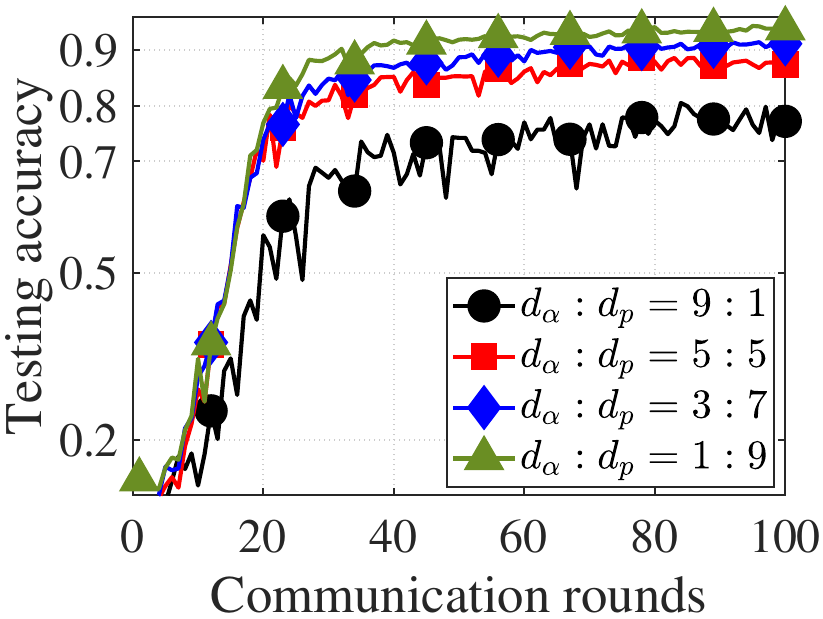}
\centerline{{(a) MNIST, WR}}\medskip
\vspace{-0.35cm}
\end{minipage}
\hspace{0.1cm}
\begin{minipage}[b]{0.48\linewidth}
\includegraphics[scale=0.50]{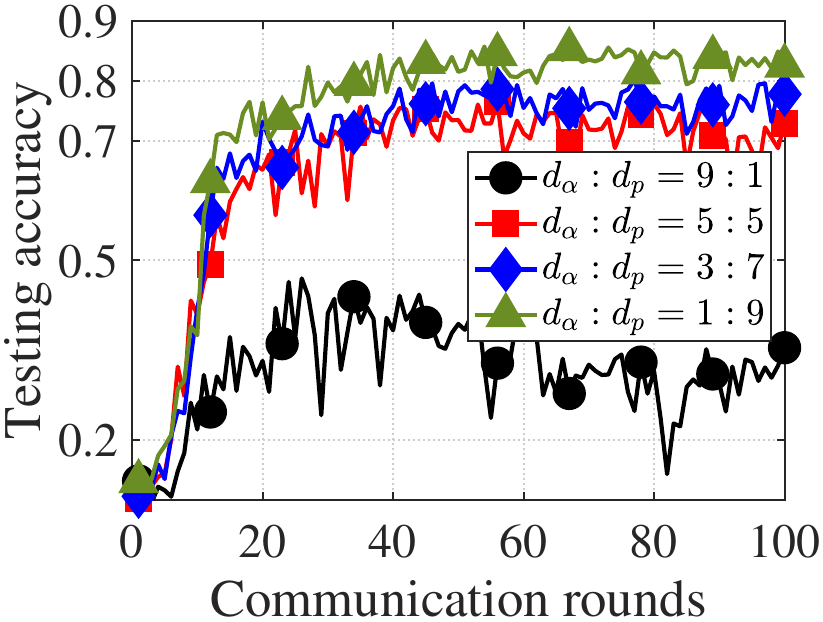}
\centerline{{(b) MNIST, WOR}}\medskip
\vspace{-0.35cm}
\end{minipage}
\caption{Testing accuracy versus communication rounds for the CNN model on the MNIST dataset under different data sampling schemes: (a) WR and (b) WOR.}
\vspace{-0.05cm}
\label{Fig:ablation_cnn}
\end{figure}

Furthermore, we conduct a supplementary CNN-based experiment on the MNIST dataset. In this experiment, we adopt a lightweight CNN consisting of two convolutional layers with ReLU activation and max-pooling, followed by two fully connected layers for classification. Specifically, the first two convolutional layers use $8$ and $16$ filters, respectively, both with kernel size $5 \times 5$, and the classifier is composed of a hidden layer with $64$ neurons and an output layer with $10$ classes. In the CNN experiment, model splitting is simulated by assigning a fixed binary mask to each layer according to the prescribed ratio $d_\alpha:d_p$, so that the masked parameters form the public submodel throughout training, while the complementary parameters are retained locally as the private submodel. The main training parameters are set as $T=100$, $Q=5$, $b=64$, and $N=20$, while the other privacy-related settings follow the corresponding experimental configuration. The results in Fig.~\ref{Fig:ablation_cnn} further corroborate the aforementioned observations. In particular, WR consistently achieves slightly better performance than WOR, and larger visible submodels still lead to faster convergence and higher testing accuracy. These results suggest that the main findings of this work are not limited to shallow models, but also extend to deeper neural network architectures.

\section{Conclusions}\label{sec:Conclusions}
In this paper, we proposed MS-PAFL, a unified framework for improving the privacy--utility trade-off in FL.  By combining structural model splitting with statistical privacy amplification, MS-PAFL partitions each client’s model into a private submodel retained locally and a public submodel used for aggregation, thereby confining DP perturbation to the released model component and reducing unnecessary distortion on the retained private part. We further developed a rigorous privacy analysis showing that random client participation and local data subsampling jointly amplify privacy protection at both the single-round and cumulative levels. Extensive experiments on non-iid data settings, together with ablation studies on the splitting ratio, validate the theoretical findings and demonstrate that MS-PAFL consistently achieves a favorable privacy--utility trade-off.

\numberwithin{equation}{section}
\begin{appendices}
\section{Proof of the theorem \ref{Thm: cdp for global model}}\label{sec:Proof of the theorem3}
Let $\overline \wb_{p}^{t}$ represent the global public model at round $t$, and $\overline \wb_{-i,p}^{t}$ be the global public model generated under the same randomness in round $t$ when client $i$ does not contribute an update (i.e., $\zeta_i^t=0$). The process satisfies $(\epsilon_c^t, \delta_c^t)$-CDP, ensuring that for any outcome $\Oc$ and any client $i$:
\begin{align} \label{eqn: cdp_A}
\!\!\!\mathbb{P}(\overline\wb_{p}^{t} \in \Oc) \leq \exp({\epsilon_{c}^t}) \cdot \mathbb{P}(\overline\wb_{-i,p}^{t} \in \Oc)+\delta_{c}^t,  \forall i \in [N].
\end{align}
Client participation in round $t$ is governed by independent check-in probabilities $p_i^t$. Consequently, the total number of participating clients, $\left|\mathcal{X}^t\right|$, can be expressed as the sum of $N$ Bernoulli random variables, with expectation $\mu_{{\Xc}}^t=\sum_{i=1}^{N} p_{i}^{t}$.

Then, we condition the left-hand side of \eqref{eqn: cdp_A} with the event that $|\Xc^{t}|$ deviates from the mean, i.e., $\left||\Xc^t| - \mu_{_{\Xc}}^t\right| \geq \beta N$ for any $\beta>0$, and bound it using Hoeffding's inequality~\cite{mohamed2021privacy}. To apply local
DP guarantee, we need additional conditioning on the event $\Ec_{i}$ that denotes the event that client $i$ participates in the FL training, i.e., $i \in \Xc^{t}$. The Hoeffding’s inequality is given by the following lemma.
\begin{Lemma}\label{Lmemma:Appendix_1}
(Hoeffding’s Inequality~\cite{mohamed2021privacy}) For a binomial random variable $|\Xc^t|$ with $N$ trials and mean $\mu_{_{\Xc}}^t$, the probability that $|\Xc^t|$ deviates from the mean by more than $\beta N$ can be bounded by,
\begin{align}
	\mathbb{P}\big(\big||\Xc^t|-\mu_{_{\Xc}}^t \big| \geq \beta N \big) \leq 2 \exp(-2 \beta^{2} N) \triangleq \delta^{\prime},  \label{Lemma: Hoeffding’s Inequality_A}
\end{align}
for any $\beta>0$ and any $\delta^{\prime} \in [0,1)$.
\end{Lemma}

Denote $p_{i}^t=\mathbb{P}(\Ec_{i}), \forall i \in [N]$. By Lemma \ref{Lmemma:Appendix_1}, the conditional probabilities $\hat{p}_{i}^t= \mathbb{P}\big(\Ec_{i} \mid \big| |\Xc^t|-\mu_{_{\Xc}}^t \big| <\beta N \big), \forall i$ can be readily bounded by total probability theorem,
\begin{align}
\mathbb{P}(\Ec_{i} ) =& \mathbb{P}\big(\Ec_{i} \mid \big||\Xc^t|-\mu_{_{\Xc}}^t \big| < \beta N\big) \mathbb{P}\big( \big||\Xc^t|-\mu_{_{\Xc}}^t \big|   < \beta N \big) \notag \\
&~+\mathbb{P}\big(\Ec_{i} \mid   \big||\Xc^t|- \mu_{_{\Xc}}^t \big| \geq \beta N \big) \mathbb{P}\big( \big||\Xc^t|-\mu_{_{\Xc}}^t \big|   \geq  \beta N \big)  \notag\\
\overset{(a)} \geq &\mathbb{P}\big(\Ec_{i} \mid   \big||\Xc^t|-\mu_{_{\Xc}}^t \big|   < \beta N\big) \mathbb{P}\big( \big||\Xc^t|-\mu_{_{\Xc}}^t \big|   < \beta N \big) \notag\\
=&\hat{p}_{i}^t (1-\delta^{\prime}), \label{eqn:full_pro_0}
\end{align}
\noindent where $(a)$ holds from the fact that any probability is lower bounded by 0.
By rearranging the two sides of \eqref{eqn:full_pro_0}, we have
\begin{align}
\hat{p}_{i}^t \leq  \frac{p_{i}^t}{ 1-\delta^{\prime} }.
\end{align}
Similarly, by total probability theorem, we have
\begin{align}
\mathbb{P}(\overline\wb_{p}^{t} \in \Oc) =&\mathbb{P}\big(\big| |\Xc^t|-\mu_{_{\Xc}}^t \big| \geq \beta N \big) \mathbb{P}\big(\overline\wb_{p}^{t} \in \Oc \mid
||\Xc^t|-\mu_{_{\Xc}}^t| \geq \beta N\big) \notag\\
&+\mathbb{P}\big( \big| |\Xc^t|-\mu_{_{\Xc}}^t \big|<\beta N \big) \mathbb{P}\big(\overline\wb_{p}^{t} \in \Oc \mid  \big| |\Xc^t|-\mu_{_{\Xc}}^t \big|  <\beta N \big) \notag\\
\overset{(a)}{\leq} & \delta^{\prime} +\mathbb{P}\big( \big| |\Xc^t|-\mu_{_{\Xc}}^t \big|<\beta N \big) \mathbb{P}\big( \overline\wb_{p}^{t} \in \Oc \mid  \big| |\Xc^t|-\mu_{_{\Xc}}^t \big| <\beta N\big), \label{eqn:Full_P_A}
\end{align}

\noindent where $(a)$ follows from the fact that any probability is upper bounded by 1. To further bound the right hand side of \eqref{eqn:Full_P_A}, we need the following lemma.
\begin{Lemma}\label{Lmemma:Appendix_21}
Let $\hat{p}_{i}^t=\mathbb{P}\big(\Ec_{i} \mid \big| |\Xc^t|-\mu_{_{\Xc}}^t \big| <\beta N\big)$. The following inequality is true when each round of local model guarantees $( \epsilon_\ell, \delta_\ell)$-LDP,
\begin{align}
	\mathbb{P}\big(\overline\wb_{p}^{t} \in \Oc \mid  \big| |\Xc^t|-\mu_{_{\Xc}}^t \big|  <\beta N\big) \leq & \Big(\hat{p}_{i}^t\big(\exp(2q_i\epsilon_\ell)-1\big)+1\Big) \mathbb{P}\big(\overline\wb_{-i,p}^{t} \in \Oc \mid   \big||\Xc^t|-\mu_{_{\Xc}}^t \big|  <\beta N \big)	\notag\\
	&+\hat{p}_{i}^t q_i\delta_\ell. \label{Lemma:bound_cdp}
\end{align}
\end{Lemma}

\vspace{0.2cm}
Then, by applying Lemma \ref{Lmemma:Appendix_21}, \eqref{eqn:Full_P_A} becomes
\begin{align}
\mathbb{P}(\overline\wb_{p}^{t} \in \Oc) \leq & \delta^{\prime}+\mathbb{P}
\big( \big| |\Xc^t|-\mu_{_{\Xc}}^t \big| < \beta N \big) \Big[\Big(\hat{p}_{i}^t\big(\exp(2q_i\epsilon_\ellb)-1\big)+1\Big) \notag\\
& \times \mathbb{P}\big(\overline\wb_{-i,p}^{t} \in \Oc \mid  \big| |\Xc^t|-\mu_{_{\Xc}}^t \big|  < \beta N \big)+\hat{p}_{i}^t q_i\delta_\ell \Big]\notag\\
\overset{(a)}{\leq} & \delta^{\prime}+\hat{p}_{i}^t q_i\delta_\ell +\mathbb{P}\big(\big| |\Xc^t|-\mu_{_{\Xc}}^t \big| < \beta N \big) \cdot\Big(\hat{p}_{i}^t\big(\exp(2q_i\epsilon_\ellb)-1\big)+1\Big)\notag\\
& \times \frac{\mathbb{P}(\overline\wb_{-i,p}^{t} \in \Oc)}{\mathbb{P}\big(\big| |\Xc^t|-\mu_{_{\Xc}}^t \big|<\beta N \big)} \notag\\
\overset{(b)}{\leq} &   \Big(\frac{p_{i}^t}{1-\delta^{\prime}}\big(\exp(2q_i\epsilon_\ellb)-1\big)+1\Big) \mathbb{P}(\overline\wb_{-i,p}^{t} \in \Oc) + \delta^{\prime}+\frac{p_{i}^t q_i}{1-\delta^{\prime}} \delta_\ell,
\end{align}where $(a)$ follows from the fact that $\mathbb{P}(\big||\Xc^t|-\mu_{_{\Xc}}^t \big| < \beta N) \leq 1$ and $\mathbb{P}(\overline\wb_{-i,p}^{t} \in \Oc,\big| |\Xc^t|-\mu_{_{\Xc}}^t \big|<\beta N)\leq \mathbb{P}(\overline\wb_{-i,p}^{t} \in \Oc)$, and $(b)$ follows from \eqref{eqn:full_pro_0}. Thus, we complete the proof. \hfill $\blacksquare$

\section{Proof of Lemma \ref{Lmemma:Appendix_21}}\label{proof of Lemma3}
With the $\Ec_{i}$ defined above, let $\Ec_{i}^{c}$ denote its complementary event. Then, by total probability theorem, we have
\begin{align}
\mathbb{P}\big(\overline\wb_{p}^{t} \in \Oc \mid  | |\Xc^t| -\mu_{_{\Xc}}^t| < \beta N \big) =&\hat{p}_{i,t} \mathbb{P}\big(\overline\wb_{p}^{t} \in \Oc\mid ||\Xc^t| -\mu_{_{\Xc}}^t| <\beta N, \Ec_{i}\big) \notag\\
	&+(1-\hat{p}_{i,t}) \mathbb{P}\big(\overline\wb_{p}^{t} \in \Oc \mid ||\Xc^t| -\mu_{_{\Xc}}^t| <\beta N, \Ec_{i}^{c}\big) \notag\\
	\stackrel{(a)}{=}&\hat{p}_{i,t} \mathbb{P}\big(\overline\wb_{p}^{t} \in \Oc \mid | |\Xc^t| -\mu_{_{\Xc}}^t| <\beta N, \Ec_{i}\big) \notag\\
	&+\big(1-\hat{p}_{i,t}\big) \mathbb{P}\big(\overline\wb_{-i,p}^{t} \in \Oc \mid ||\Xc^t| -\mu_{_{\Xc}}^t| <\beta N\big),
\end{align}
\noindent where we can show that $(a)$ is true as follows:
\begin{align}\label{eqn:B2}
	\mathbb{P}\big(\overline\wb_{p}^{t} \in \Oc\mid | |\Xc^t| -\mu_{_{\Xc}}^t|   <\beta N, \Ec_{i}^{c}\big) =&\sum_{A_{-i}^t \subseteq [N], \atop | | A_{-i}^t| -\mu_{_{\Xc}}^t |<\beta N} \mathbb{P}\big(|\Xc^t|=A_{-i}^t\mid ||\Xc^t| -\mu_{_{\Xc}}^t| <\beta N, \Ec_{i}^{c}\big)\notag\\
	&\times \mathbb{P}\big(\overline\wb_{p}^{t} \in \Oc\mid ||\Xc^t| -\mu_{_{\Xc}}^t| <\beta N, \Ec_{i}^{c}, |\Xc^t|=A_{-i}^t\big) \notag\\
	\stackrel{(a)}{=} &\sum_{A_{-i}^t \subseteq[N], \atop || A_{-i}^t| - \mu_{_{\Xc}}^t| <  \beta N} \mathbb{P}\big(|\Xc^t|=A_{-i}^t \mid ||\Xc^t| -\mu_{_{\Xc}}^t| <\beta N\big) \notag\\
	&\times \mathbb{P}\big(\overline\wb_{p}^{t} \in \Oc \mid ||\Xc^t| -\mu_{_{\Xc}}^t| <\beta N, \Ec_{i}^{c}, |\Xc^t|=A_{-i}^t\big) \notag\\
	\stackrel{(b)}{=} &\sum_{A_{-i}^t \subseteq[N], \atop || A_{-i}^t| - \mu_{_{\Xc}}^t| <  \beta N}  \mathbb{P}\big(|\Xc^t|=A_{-i}^t \mid ||\Xc^t| -\mu_{_{\Xc}}^t| <\beta N\big) \notag\\
	&\times \mathbb{P} \big(\overline\wb_{-i,p}^{t} \in \Oc \mid ||\Xc^t| -\mu_{_{\Xc}}^t| <\beta N, |\Xc^t|=A_{-i}^t\big) \notag\\
	=&\mathbb{P}\big(\overline\wb_{-i,p}^{t} \in \Oc\mid ||\Xc^t| -\mu_{_{\Xc}}^t| <\beta N\big)
\end{align}
\noindent where $(a)$ holds since client $i$ is not in the set $A_{-i}^t$. Therefore, conditioning on the event $\Ec_{i}^{c}$ does not change the probability; and $(b)$ follows due to similar argument. Next, we upper bound $\mathbb{P}\big(\overline\wb_{p}^{t} \in \Oc \mid ||\Xc^t|-\mu_{_{\Xc}}^t| <\beta N, \Ec_{i}\big)$ as follows:
\begin{align}
	\mathbb{P}\big(\overline\wb_{p}^{t} \in \Oc\mid ||\Xc^t| -\mu_{_{\Xc}}^t| <\beta N, \Ec_{i}\big) = &\sum_{A^t \subseteq[N]: i \in A^t, \atop ||A^t|-\mu_{_{\Xc}}^t \mid<\beta N} \mathbb{P}\big(|\Xc^t|=A^t \mid ||\Xc^t| -\mu_{_{\Xc}}^t| <\beta N, \Ec_{i}\big) \notag\\
	&\times \mathbb{P}\big(\overline\wb_{p}^{t} \in \Oc\mid ||\Xc^t| -\mu_{_{\Xc}}^t| <\beta N, \Ec_{i}, |\Xc^t|=A^t\big).\label{lemma4-3}
\end{align}

\noindent According to privacy amplification via subsampling (cf. Theorem~\ref{thm: privacy amplicfication via sampling}), if each client applies a mechanism that satisfies $(\epsilon_{\ell}, \delta_{\ell})$-LDP, then under a subsampling ratio $q_i$, the mechanism in fact guarantees $(2q_i\epsilon_{\ell}, q_i\delta_{\ell})$-LDP. Therefore, we have
\begin{align}
	&\mathbb{P}\big(\overline\wb_{p}^{t} \in \Oc\mid ||\Xc^t| -\mu_{_{\Xc}}^t| <\beta N
	, \Ec_{i},  |\Xc^t|=A^t\big)\notag\\
	\leq & \exp{(2q_i\epsilon_{\ell})} \mathbb{P}\big(\overline\wb_{p}^{t} \in \Oc\mid ||\Xc^t| -\mu_{_{\Xc}}^t| <\beta N, \Ec_{i}^{c},  |\Xc^t|=A_{-i}^t\big) + q_i\delta_{\ell}.\label{lemma4-4}
\end{align}
Plugging \eqref{lemma4-4} into \eqref{lemma4-3}, we obtain the following:
\begin{align}
\mathbb{P}\big(\overline\wb_{p}^{t} \in \Oc \mid ||\Xc^t| -\mu_{_{\Xc}}^t| <\beta N, \Ec_{i}\big) \leq & \sum_{A^t \subseteq[N]: i \in A^t, \atop | |A^t| -\mu_{_{\Xc}}^t|<\beta N}
	\mathbb{P}\big( |\Xc^t|=A^t\mid ||\Xc^t| -\mu_{_{\Xc}}^t| <\beta N, \Ec_{i}\big)\Big(\exp(2q_i\epsilon_{\ellb}) \notag\\
	& \cdot\mathbb{P}\big(\overline\wb_{p}^{t} \in \Oc\mid ||\Xc^t| -\mu_{_{\Xc}}^t| <\beta N, \Ec_{i}^{c},  |\Xc^t|=A_{-i}^t\big) + q_i\delta_{\ellb}\Big) \notag\\
	\stackrel{(a)}{\leq}&\sum_{A^t \subseteq[N]: i \in A^t, \atop | |A^t| -\mu_{_{\Xc}}^t|<\beta N} \mathbb{P}\big( |\Xc^t|=A^t \mid ||\Xc^t| -\mu_{_{\Xc}}^t| <\beta N, \Ec_{i} \big) \cdot\exp(2q_i\epsilon_{\ellb})\notag\\
&\cdot \mathbb{P}\big(\overline\wb_{p}^{t} \in \Oc\mid ||\Xc^t| -\mu_{_{\Xc}}^t| <\beta N, \Ec_{i}^{c},  |\Xc^t|=A_{-i}^t\big)+ q_i\delta_{\ellb} \notag\\
	\stackrel{(b)}= &\sum_{A^t \subseteq[N]: i \in A^t, \atop | |A^t| -\mu_{_{\Xc}}^t|<\beta N} \mathbb{P}\big( |\Xc^t|=A_{-i}^t \mid ||\Xc^t| -\mu_{_{\Xc}}^t| <\beta N\big) \cdot \exp(2q_i\epsilon_{\ellb})\notag\\
	&\cdot \mathbb{P}\big(\overline\wb_{p}^{t} \in \Oc\mid ||\Xc^t| -\mu_{_{\Xc}}^t| <\beta N, \Ec_{i}^{c},  |\Xc^t|=A_{-i}^t\big) + q_i \delta_{\ellb} \notag\\
	\stackrel{(c)}=&\sum_{A^t \subseteq[N]: i \in A^t, \atop | |A^t| -\mu_{_{\Xc}}^t|<\beta N} \mathbb{P}\big( |\Xc^t|=A_{-i}^t\mid ||\Xc^t| -\mu_{_{\Xc}}^t| <\beta N\big) \cdot \exp(2q_i\epsilon_{\ellb})  \notag\\
	&\cdot\mathbb{P}\big(\overline\wb_{-i,p}^{t} \in \Oc\mid ||\Xc^t| -\mu_{_{\Xc}}^t| <\beta N,  |\Xc^t|=A_{-i}^t\big) + q_i\delta_{\ellb} \notag\\
	=&\exp(2q_i\epsilon_{\ellb}) \mathbb{P}\big(\overline\wb_{-i,p}^{t} \in \Oc \mid ||\Xc^t| -\mu_{_{\Xc}}^t| <\beta N\big)+ q_i\delta_{\ellb},
\end{align}

\noindent where $(a)$ holds from the fact that any probability is upper bounded by 1, $(b)$ and $(c)$ follow from arguments similar to the one used in \eqref{eqn:B2}.
\begin{align}
	\mathbb{P}\big(\overline\wb_{p}^{t} \in \Oc\mid ||\Xc^t| -\mu_{_{\Xc}}^t| <\beta N\big) \leq& \hat{p}_{i,t} \exp(2q_i\epsilon_{\ell}) \mathbb{P}\big(\overline\wb_{-i,p}^{t} \in \Oc \mid ||\Xc^t| -\mu_{_{\Xc}}^t| <\beta N\big) \notag\\
	&+\hat{p}_{i,t} q_i\delta_{\ell}+(1-\hat{p}_{i,t}) \mathbb{P}\big(\overline\wb_{-i,p}^{t} \in \Oc \mid ||\Xc^t| - \mu_{_{\Xc}}^t| <\beta N\big).
\end{align}
\noindent Rearranging the above inequality, we can obtain the result of Lemma \ref{Lmemma:Appendix_21}. \hfill $\blacksquare$

\section{Proof of Theorem \ref{Thm:total_budget}} \label{subsec:the_Proof_of_Theorem_dp}
The proof mainly follows the general approach of~\cite{li2025differentially}, while explicitly incorporating both data subsampling and random client participation. To proceed, let $\mathbf{u}_{i}$ denote the unique data sample that differs between neighboring datasets $\mathcal{D}_{i}$ and $\mathcal{D}_{i}^{\prime}$. We then define the $\lambda$-th log moment of the privacy loss for client $i$ at the $t$-th communication round as follows:
\begin{align}\label{eqn:log_moment}
\mathcal{L}_{i}^{t}(\lambda) =&   \ln \Big( \mathbb{E}_{\widetilde{{\wb}}_{i , p}^{t}} \Big[ \Big( \frac{\operatorname{Pr}\big[\widetilde{{\wb}}_{i , p}^{t} \mid \mathcal{D}_{i}\big]}{\operatorname{Pr}\big[\widetilde{{\wb}}_{i , p}^{t} \mid \mathcal{D}_{i}^{\prime}\big]} \Big)^{\lambda} \Big] \Big).
\end{align}
According to \cite[Theorem~2]{abadi2016deep}, when each client guarantees $(\epsilon_\ell, \delta_\ell)$-LDP, the $\mathcal{L}_{i}^{t}(\lambda)$ is given by
\begin{align}\label{eqn:log_moment_1}
\mathcal{L}_{i}^{t}(\lambda) =  \frac{q_i^{2}}{1-q_i}\cdot \frac{\lambda(\lambda + 1) \epsilon_{\ell}^{2}}{4 s^2 \ln (1.25 / \delta_\ell)},
\end{align}
where $s$ denotes the $\ell_2$ sensitivity of public submodel $\wb_{i,p}^t$, as defined in \eqref{eqn:global sensitivity_f}. For the MS-PAFL framework, this sensitivity is bounded as:
\begin{align}\label{eqn:global sensitivity_f_appendix}
s = \max _{\mathcal{D}_i, \mathcal{D}_i^{\prime}}\big\|\wb_{i,p}^t (\mathcal{D}_i)- \wb_{i,p}^t \big(\mathcal{D}_i^{\prime}\big)\big\| \overset{(a)}{\leq} 2 \mathcal{C}.
\end{align}where $(a)$ is a direct result of applying gradient clipping with threshold $\mathcal{C}$ during the local update, as specified in \eqref{eqn:clipping}.

Then, by substituting \eqref{eqn:global sensitivity_f_appendix} into \eqref{eqn:log_moment_1} and applying Theorem~2 (linear composability property) in \cite{abadi2016deep}, the log moment of the total privacy loss of client $i$ after $T$ rounds can be obtained as follows.
{\small\begin{align} \label{eqn:total_momentaccount}
\mathcal{L}_{i}(\lambda) &=  \sum_{t=1}^{T} \mathbb{P} (i \in \Xc_t)\mathcal{L}_{i}^{t}(\lambda) =   \frac{ p_i T q_i^{2}  \epsilon_{\ell}^{2} \cdot\lambda(\lambda+1)}{16 \mathcal{C}^2 (1-q_i) \ln (1.25 / \delta_\ell)},
\end{align}}where $p_i$ denotes the check-in probability of client $i$.

Since the local clients ensure $(\epsilon_\ell, \delta_\ell)$-LDP, by Theorem 2 in \cite{abadi2016deep}, we have
\begin{align}\label{eqn:tail_bound}
\ln (\delta_\ell) &=\min _{\lambda \in \mathbb{Z}^{+}} \big(\mathcal{L}_{i}(\lambda) - \lambda \bar{\epsilon}_{i}^T \big) \notag\\
&=\min _{\lambda \in \mathbb{Z}^{+}} \big(  \frac{p_i T q_i^{2}  \lambda(\lambda+1) \epsilon_{\ell}^{2}}{16 \mathcal{C}^2 (1-q_i) \ln (1.25 / \delta_\ell)}- \lambda \bar{\epsilon}_{i}^T \big).
\end{align}
Since problem~\eqref{eqn:tail_bound} is a quadratic function with respect to $\lambda$, by  plugging its minimal value into \eqref{eqn:tail_bound}, we have
\begin{align}
\ln(\delta_\ell) = & \min _{\lambda  \in \mathbb{Z}^{+}} \big(  \frac{p_i T q_i^{2}  \lambda(\lambda + 1) \epsilon_{\ell}^{2}}{16 \mathcal{C}^2 (1-q_{i}) \ln (1.25 / \delta_\ell)}- \lambda \bar{\epsilon}_{i}^T \big) \notag \\
\geq & \frac{3 p_i T q_{i}^{2}   \epsilon_{\ell}^{2}}{64 \mathcal{C}^2 (1-q_{i}) \ln (1.25 / \delta_\ell)}  - \frac{(\bar{\epsilon}_{i}^T)^{2} 4\mathcal{C}^2 (1-q_{i}) \ln (1.25 / \delta_\ell)}{p_i T q_{i}^{2}   \epsilon_{\ell}^{2}}, \label{eqn:minlntau}
\end{align}
which further implies
\begin{align}
\ln(1/\delta_\ell)  \leq  \frac{(\bar{\epsilon}_{i}^T)^{2} 4\mathcal{C}^2 (1-q_{i}) \ln (1.25 / \delta_\ell)}{p_i T q_{i}^{2}   \epsilon_{\ell}^{2}}. \label{eqn:minlntau_inverse}
\end{align}
By rearranging \eqref{eqn:minlntau_inverse}, we have
\begin{align}
\bar{\epsilon}_{i}^T \geq   \frac{q_i}{2\mathcal{C}  \sqrt{1-q_i}} \sqrt{ \frac{p_i T  \ln(1/\delta_\ell)}{ \ln (1.25 / \delta_\ell)}} \epsilon_\ell.
\end{align}

\noindent Therefore, there exists a constant $ c_{0}$ such that the total privacy loss of client $i$ after $T$ rounds satisfies:
\begin{align} \label{eqn:total_privacyloss_appdix}
\bar{\epsilon}_{i}^T =c_{0} \frac{q_i \sqrt{p_i}}{\sqrt{1-q_i}} \sqrt{T} \epsilon_\ell, \ \forall  i \in [N].
\end{align}
Thus, we complete the proof. $\hfill\blacksquare$
\end{appendices}

\bibliographystyle{IEEEtran}
\bibliography{reference}

\end{document}